\documentclass[11pt]{article}

\usepackage[margin=1in]{geometry}
\usepackage[T1]{fontenc}
\usepackage[utf8]{inputenc}
\usepackage{lmodern}
\usepackage{microtype}
\usepackage{amsmath,amssymb,mathtools}
\usepackage{booktabs}
\usepackage{graphicx}
\usepackage{multirow}
\usepackage{array}
\usepackage{xcolor}
\usepackage{natbib}
\usepackage[hidelinks]{hyperref}
\usepackage[nameinlink,noabbrev]{cleveref}
\usepackage{placeins}
\usepackage{longtable}
\usepackage{algorithm}
\usepackage{algpseudocode}
\usepackage{amsthm}
\usepackage{enumitem}


\theoremstyle{plain}
\newtheorem{theorem}{Theorem}

\theoremstyle{definition}
\newtheorem{assumption}{Assumption}

\theoremstyle{plain}

\graphicspath{{overleaf_figures/}}

\newcommand{\Pp}{\mathbb{P}}

\newcommand{\muhat}{\widehat\mu}

\title{TabCUQ: Context-Uncertainty Quantification for Tabular Foundation Models\thanks{Correspondence to:
    Molei Liu (\texttt{moleiliu@bjmu.edu.cn}) and
    Zijian Guo (\texttt{zijguo@zju.edu.cn}).}}
\author{
    Yuebei Shi$^{1,2}$,
    Molei Liu$^{3,4}$
    and Zijian Guo$^{2}$\\[0.5em]
    {\small $^{1}$School of Mathematics, Zhejiang University}\\ {\small $^{2}$Center for Data Science, Zhejiang University}\\
    {\small $^{3}$Department of Biostatistics, Peking University Health Science Center}\\
    {\small $^{4}$Beijing International Center for Mathematical Research, Peking University}\\
}
\date{}

\begin{document}
\maketitle

\begin{abstract}
\noindent {Tabular foundation models use in-context learning to estimate conditional expected outcomes and average treatment effects for specific groups of individuals. Using these estimates to guide scientific decisions requires reliable uncertainty quantification. The challenge is to account simultaneously for two sources of uncertainty: limited-context bias and context variation arising from the small labeled sample supplied at inference. Predictive spread does not directly quantify both sources of uncertainty, and we observe that the ordinary bootstrap may substantially overestimate context variation, whereas subsampling may underestimate it. We introduce TabCUQ, a perturbation-based framework that injects noise into observed outcomes and applies the foundation model to the perturbed contexts. TabCUQ estimates context variation using wild-bootstrap dispersion and accounts for limited-context bias using the maximum deviation between perturbed and original estimates. For a kernel-regression analogue whose estimation error admits an exact decomposition into these two components, we show that our framework provides asymptotically valid pointwise coverage. In simulations, TabCUQ substantially improves coverage, with average interval widths typically about twice those of an oracle bias-aware benchmark. Across seven real-world tasks, it also improves coverage of reference functions learned from much larger samples.}

\end{abstract}

\section{Introduction}

Scientific decisions often depend on estimates drawn from small tabular datasets. Tabular foundation models offer a promising approach to prediction in these settings. Conventional methods, such as generalised additive models, random forests and neural networks trained from scratch, are typically fitted separately to each dataset \citep{hastie2009elements}. The foundation models studied here are neural networks trained in advance on many synthetic and real datasets. This training allows the same model to be reused for prediction tasks on new datasets. For a new task, the user supplies a small labeled dataset containing examples of features and their observed outcomes. The model takes these examples as part of its input and uses them to predict the outcome of a new observation from its features, without updating its pretrained parameters. This process is known as in-context learning. Built on this idea, tabular foundation models, including TabPFN \citep{hollmann2025tabpfn}, TabICL \citep{qu2025tabicl} and TabFM \citep{kong2026tabfm}, have demonstrated strong predictive performance on small-data benchmarks. Their predictive success raises a question: how reliably can estimates obtained from a small labeled dataset support a scientific conclusion?

We focus on two quantities that inform scientific decisions: the conditional mean and the conditional average treatment effect. A chemist may need to determine whether the mean reaction yield under specified experimental conditions exceeds a threshold for further development. Time and material costs limit the number of experiments available to estimate this quantity \citep{shields2021bayesian,hickman2022constraints}. The average outcome across repeated experiments under those conditions is a \emph{conditional mean}. A clinical researcher may instead ask whether a treatment shortens average recovery time for patients with a particular clinical profile. Their average recovery time under treatment minus their average recovery time without treatment is the \emph{conditional average treatment effect} (CATE); a negative value indicates faster recovery. 

\begin{figure}[htp!]
    \centering
    \includegraphics[width=\linewidth]{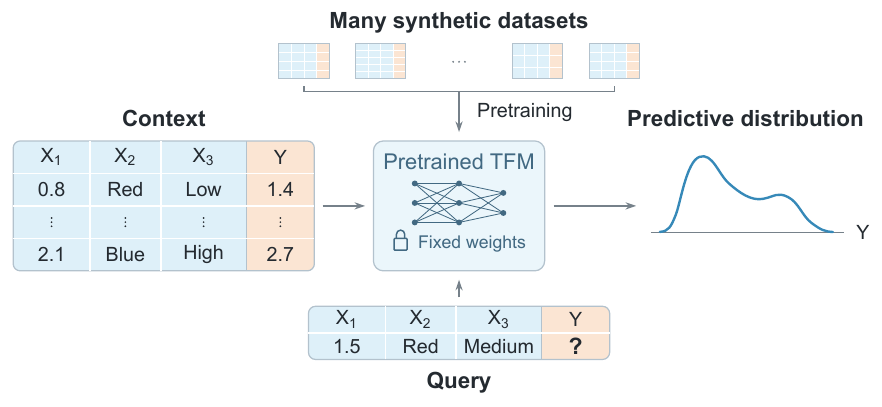}
    \caption{In-context prediction with a tabular foundation model (TFM) pretrained on many synthetic datasets. The context contains observed feature--outcome pairs, while the query supplies feature values with its outcome $Y$ left unknown. With its pretrained weights fixed, the model processes both inputs to produce a predictive distribution for $Y$.}
    \label{fig:in-context-conditional-mean}
\end{figure}

We use conditional-mean estimation to explain how a foundation model produces an estimate. As illustrated in Figure~\ref{fig:in-context-conditional-mean}, the foundation model takes two inputs: context and query. The \emph{context} is the labeled dataset containing observed features and outcomes, and the \emph{query} specifies the feature values at which the mean is to be estimated. In the chemistry example, the context contains completed experiments and their measured yields, while the query specifies the reaction conditions of interest. The model processes the context and query without updating its pretrained weights, a procedure called \emph{in-context learning}. We use its point prediction as an estimator of the conditional mean. Although the prediction rule is fixed, the estimate depends on the labeled observations supplied as context. 

\begin{figure}[!htb]
    \centering
    \includegraphics[width=\linewidth]{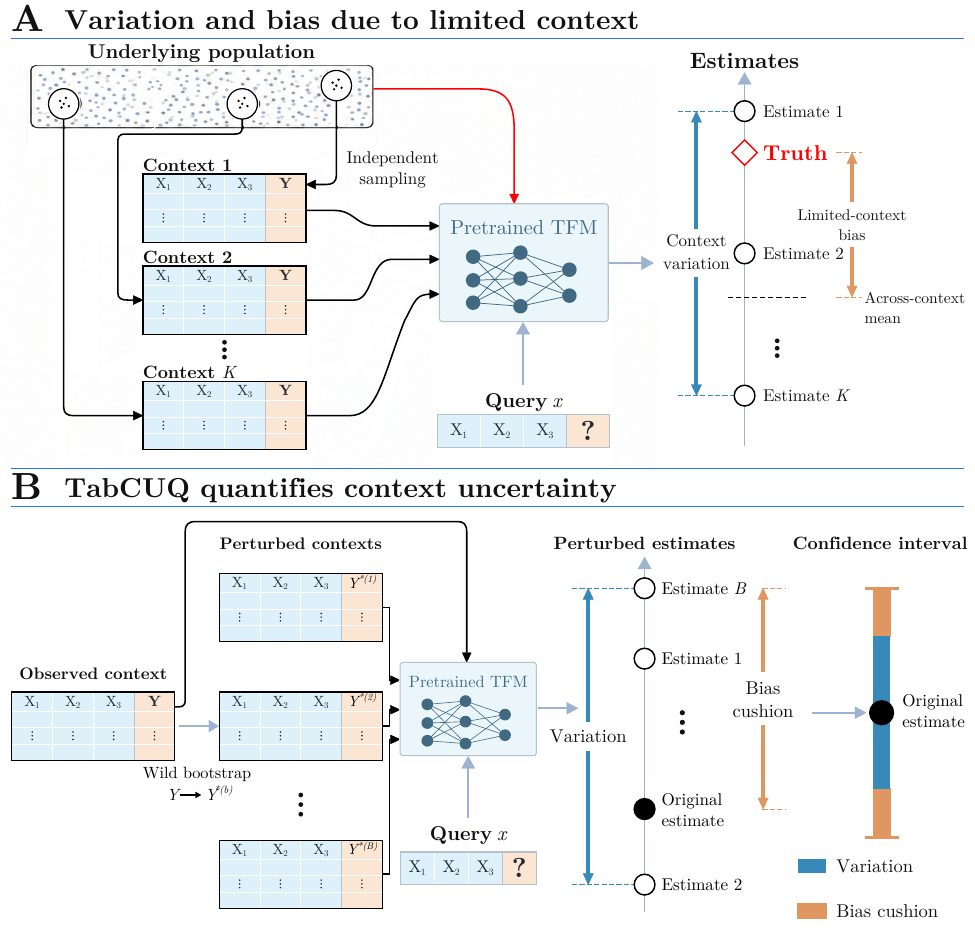}
    \caption{\textbf{TabCUQ quantifies context uncertainty.}
    \textbf{(A)} With the pretrained TFM and query fixed, independently sampled contexts yield different estimates. Their spread represents context variation, while the gap between their across-context mean and truth represents limited-context bias.
    \textbf{(B)} Wild-bootstrap outcome perturbations provide an estimate of context variation and an empirical bias cushion. The blue central segment represents the contribution of estimated variation, while the orange extensions represent the empirical bias cushion.}
    \label{fig:overview}
\end{figure}

We distinguish two components of \emph{context uncertainty}: context variation and limited-context bias. A fresh context of the same size sampled from the target population can produce a different estimate. Each context consists of newly sampled observations, so both features and outcomes vary. We refer to this sampling variation as \emph{context variation}. To understand the second component, imagine repeatedly drawing contexts of the same size and averaging the resulting estimates at a fixed query. Even after the sampling fluctuations are averaged out, the average estimate may still lie above or below the true conditional mean at that query. We refer to this systematic gap as \emph{limited-context bias}. { In our experiments, TabPFN frequently exhibits such bias with small contexts, analogous to the regularization bias commonly encountered in classic machine learning methods.} In statistical terminology, these are the sampling variability and bias of the estimator induced by the pretrained prediction rule. The same distinction applies to a CATE estimator at a fixed clinical profile. Panel A of \cref{fig:overview} illustrates both components. Both components matter when scientific decisions depend on an estimated mean or treatment effect. Sampling fluctuations or bias can make an estimate appear favorable even when the true mean yield is inadequate or the treatment benefit is negligible. A point estimate alone therefore does not establish whether the data support the scientific conclusion. We seek confidence intervals for the conditional mean or CATE. A valid confidence-interval procedure covers the fixed target with probability at least the stated confidence level over repeated contexts of the same size. Such intervals help assess whether the data support a mean yield above the development threshold or a treatment effect below zero, taking estimation uncertainty into account.

Constructing such confidence intervals requires estimating context variation and accounting for limited-context bias. Bootstrap and subsampling provide standard tools for approximating sampling distributions under appropriate conditions and are natural candidates for estimating context variation. Applying classical resampling methods to pretrained tabular foundation models raises the question of whether they accurately capture variation across the labeled contexts supplied at inference, while the pretrained network parameters remain fixed. Nevertheless, accurately estimating context variation is only part of the inferential task. In nonparametric regression, estimation bias that is non-negligible relative to the estimator's standard deviation can impair confidence-interval coverage, even when that standard deviation is accurately estimated \cite[Section~5.7]{wasserman2006all}. Pretraining does not itself guarantee negligible bias at the available context size. Through extensive simulation and real-world studies, we observe that the Tabular foundation models tend to suffer from an estimation bias with the limited aomunt of context. We therefore develop TabCUQ, a perturbation-based confidence-interval construction that accounts for both context variation and limited-context bias.

\subsection{Main results and contributions}
The paper makes two main contributions. First, we find empirically that wild-bootstrap perturbations effectively estimate the scale of context variation for the pretrained tabular prediction rules evaluated here. Interestingly, we observe that the ordinary bootstrap produces standard-error estimates substantially larger than the true standard deviations, whereas the evaluated subsampling procedure produces smaller estimates. These findings motivate our use of wild-bootstrap perturbations in the uncertainty quantification of the tabular foundation model. Second, we introduce TabCUQ, a perturbation-based confidence-interval construction for conditional means and CATEs that combines an estimate of context variation with a characterization of limited-context bias. In a kernel-regression setting, we prove that the proposed bias cushion bounds estimation bias and establish coverage for the resulting interval under the stated assumptions.

TabCUQ generates perturbed contexts by applying wild-bootstrap perturbations to the observed outcomes, while holding the features and context size fixed. We then apply the tabular foundation model to each perturbed context to obtain the corresponding perturbed estimates. We use the empirical standard deviation of these estimates to estimate the scale of context variation. To characterize limited-context bias, we define a \emph{bias cushion} as the largest absolute displacement of the perturbed estimates from the original estimate. We construct a normal-approximation confidence interval centered on the original point estimate and widen it by this cushion on each side. The construction thus uses the same perturbed estimates both to quantify sampling variability and to provide an additional margin for bias.

{We evaluate TabCUQ in extensive simulated tasks of conditional-mean and CATE inference with three tabular foundation models. Even when wild-bootstrap standard errors are reasonably well calibrated, confidence intervals that ignore limited-context bias can substantially undercover. TabCUQ improves coverage across the main simulation settings, generally reaching the nominal 95\% level, with average interval widths typically around twice those of an oracle benchmark that uses the true bias and variance. We further evaluate TabCUQ on seven real-world tasks spanning socioeconomic, environmental, behavioural, protein and single-cell data. Since the true conditional means are unknown, we assess coverage relative to reference functions fitted on much larger samples. With contexts of small sizes, the average reference coverage across tasks increases from 0.614–0.842 for wild bootstrap intervals to 0.966–1.000 for TabCUQ, while average interval widths of TabCUQ remain around two times those of the oracle benchmark.}

\subsection{Relation to existing work}

TabCUQ builds on classical resampling methods for inference about fixed population quantities. The nonparametric bootstrap and subsampling approximate sampling distributions by recomputing estimators on resampled observations \citep{efron1979bootstrap,politis1999subsampling}. Classical wild-bootstrap methods perturb regression residuals with covariates held fixed to accommodate heteroskedastic errors \citep{wu1986jackknife}. TabCUQ applies outcome perturbations to pretrained in-context prediction rules, using the resulting estimates to quantify context variation and construct a margin for limited-context bias. Its use of a bias cushion also connects the construction to bias-aware methods that calibrate confidence intervals to accommodate estimation bias \citep{armstrong2016honest}.

For pretrained models, posterior methods offer another approach to uncertainty quantification. PFNs were introduced to approximate Bayesian posterior predictive distributions under a specified prior over tasks \citep{muller2022transformers}. Predictive intervals from these distributions concern future responses and include outcome noise. Posterior methods can also target uncertainty about conditional means and other functionals. The approximate martingale posterior method of \citet{nagler2025uncertainty} uses PFN-initialised predictive distributions and copula updates to quantify uncertainty about conditional means, quantiles and other summaries at a fixed query. The authors motivate learning-rate choices that widen credible intervals to accommodate nonparametric bias. TabMGP \citep{ng2026tabmgp} constructs martingale posteriors for parameters defined as minimisers of an expected loss. \citet{fortini2026uncertainty} use a predictive central limit theorem to obtain Gaussian approximations to posterior uncertainty about conditional event probabilities under a limiting predictive distribution. 
These approaches provide posterior uncertainty for a range of targets using pretrained models. In contrast, our proposal assesses the uncertainty from the frequentist perspective, {which has distinct properties and different application scenarios from posterior credibility.} Our method constructs intervals by perturbing the observed outcomes and repeatedly evaluating the original point estimator. We assess their coverage at a fixed query over repeated draws of the labeled context. 

{Conformal prediction constructs prediction sets for future responses \citep{shafer2008tutorial}, whereas TabCUQ constructs confidence intervals for conditional means and CATEs. Under exchangeability, standard split conformal prediction guarantees marginal coverage \citep{lei2018distribution}; coverage at a particular query can still fall below the nominal level \citep{barber2021limits}.} For tabular foundation models, C-USIM \citep{park2026conformal} uses held-out observations to calibrate highest predictive density regions and studies the gap between conditional and marginal coverage. In causal inference, \citet{lei2021conformal} develop conformal methods for counterfactual outcomes and individual treatment effects. An individual treatment effect is the difference between a person's two potential outcomes; CATE is the conditional expectation of this difference given the features. These are different inferential targets: even valid coverage for outcomes or individual effects at a fixed query does not by itself establish confidence-interval coverage for their conditional means.

\section{Uncertainty Quantification for Tabular Foundation Model}
\label{sec:setup}

Uncertainty quantification is essential when predictions from tabular foundation models inform decisions based on small- to medium-sized datasets. A positive estimate of a conditional mean or treatment effect does not, by itself, establish that the underlying quantity is positive. The estimated sign may reflect limited-context bias or context variation. We first review tabular foundation models in Section~\ref{sec: basics} and then discuss the need for uncertainty quantification and the challenges involved in the remainder of this section.

\subsection{Basics of Tabular Foundation Models}
\label{sec: basics}

We review prior-data fitted networks (PFNs), which provide the methodological foundation for TabPFN \citep{muller2022transformers}. PFNs are pretrained on synthetic datasets and subsequently applied to new datasets at prediction time. Let $\phi$ index a data-generating rule, with prior density $\pi(\phi)$. To construct each synthetic dataset, the simulator first samples $\phi\sim\pi$ and then generates labeled observations independently from the sampled rule, conditional on $\phi$. The observations are randomly divided into a labeled context and held-out queries. For each query, the network receives the context together with the query features and predicts the corresponding outcome label. The query labels are withheld from the network inputs and used only to compute the training loss.

Let $D$ denote the labeled context and $(X_{\mathrm{test}},Y_{\mathrm{test}})$ a held-out query, where $X_{\mathrm{test}}$ and $Y_{\mathrm{test}}$ are its features and outcome label, respectively. A PFN with parameters $\theta$ outputs a predictive density $q_\theta(y\mid x,D)$ for the outcome label of a query with features $x$, given context $D$. The population pretraining objective is

\begin{equation}
    \mathcal{L}(\theta)
    =
    \mathbb{E}_{\phi\sim\pi}
    \mathbb{E}_{D,(X_{\mathrm{test}},Y_{\mathrm{test}})\mid\phi}
    \left[
        -\log q_\theta(Y_{\mathrm{test}}\mid X_{\mathrm{test}},D)
    \right].
    \label{eq:pfn-loss}
\end{equation}
In practice, we sample synthetic datasets using the procedure described above and approximate the expectation in \eqref{eq:pfn-loss} by an empirical average of the losses on the sampled synthetic datasets. This empirical loss is used to update the network parameters $\theta$, with new synthetic datasets sampled throughout training.

We follow \citet{muller2022transformers} and connect the pretraining objective in \eqref{eq:pfn-loss} to the Bayesian posterior predictive density. Let $p_{\pi}(y\mid x,D)$ denote the Bayesian posterior predictive density of the query outcome with the prior $\pi$, admitting the following expression \begin{equation}
    p_{\pi}(y\mid x,D)
    =
    \int p(y\mid x,\phi)\,
         p_{\pi}(\phi\mid x,D)\,d\phi.
    \label{eq:pfn-posterior-predictive}
\end{equation}
{Here, $p(y\mid x,\phi)$ is the conditional response density under rule $\phi$, and $p_{\pi}(\phi\mid x,D)$ is the posterior density of the rule given the context and query features. 
The loss in \eqref{eq:pfn-loss} decomposes as
\begin{equation}
    \mathcal{L}(\theta)
    =
    C
    +
    \mathbb{E}_{D,X_{\mathrm{test}}}
    \left[
        \mathrm{KL}\!\left(
            p_{\pi}(\cdot\mid X_{\mathrm{test}},D)
            \,\Vert\,
            q_\theta(\cdot\mid X_{\mathrm{test}},D)
        \right)
    \right],
    \label{eq:pfn-loss-kl}
\end{equation}
where $C=\mathbb{E}[-\log p(Y_{\mathrm{test}}\mid X_{\mathrm{test}},D)]$ is independent of $\theta$. Minimizing the population loss is therefore equivalent to minimizing the expected KL divergence from $p_{\pi}$. Over a rich class of $q_{\theta}$ functions, this divergence is minimized with $q_{\theta}(\cdot\mid x,D)=p_{\pi}(\cdot\mid x,D)$ for almost every $(D,x)$. Thus, $p_{\pi}$ is the population predictive target, although a fitted PFN may not achieve it due to restrictions on the network class, finite pretraining data, and imperfect optimization.

Conventional Bayesian methods typically obtain the predictive density by approximating this posterior distribution $p_{\pi}(\phi\mid x,D)$ for each context and then applying \eqref{eq:pfn-posterior-predictive} to learn the posterior outcome distribution. In contrast, PFNs are based on directly using $q_\theta(y\mid x,D)$ to approximate the posterior predictive density $p_{\pi}(y\mid x,D)$ through pretraining on simulated datasets, without explicitly constructing the posterior over $\phi$ or evaluating the integral in \eqref{eq:pfn-posterior-predictive} at prediction time.}

\subsection{Inferential Targets and Real-data Examples}

We consider two inferential targets that inform the scientific decisions. 
The first target is the conditional mean in regression, which describes the average outcome for given covariates. Let $D_n=\{(X_i,Y_i)\}_{i=1}^n$ be a context consisting of i.i.d.\ observations, and fix a query feature vector $x$. Our target is the population conditional mean $\mu(x)=\mathbb E[Y\mid X=x]$. The tabular foundation model provides a predictive density $q_\theta(y\mid x,D_n)$ and the corresponding conditional mean estimator
\begin{equation}
    \widehat\mu(x)
    =
    \int y\,q_\theta(y\mid x,D_n)\,dy.
    \label{eq:pfn-mean-estimator}
\end{equation}

The second target is the conditional average treatment effect (CATE), defined as the difference between the conditional means of the potential outcomes under treatment and control \citep{athey2016recursive}. Let $A\in\{0,1\}$ indicate treatment, $X$ denote pretreatment covariates, and $Y(a)$ denote the potential outcome under treatment assignment $a$. At a given covariate value $x$, we define the corresponding CATE as $\tau(x)=\mathbb E[Y(1)-Y(0)\mid X=x]$. To identify $\tau(x)$ from observed data, we assume the standard causal identification condition:  (i) consistency, $Y=Y(A)$; (ii) overlap, $0<P(A=1\mid X=x)<1$ for all possible $x$;  (iii)unconfoundedness, $\{Y(0),Y(1)\}\perp A\mid X$ \citep{hernan2020causal}. Writing $\mu_a(x)=\mathbb E[Y\mid A=a,X=x]$ for $a\in\{0,1\}$, these assumptions give $\tau(x)=\mu_1(x)-\mu_0(x)$. This representation reduces CATE estimation to two regression problems. Given a context $D_n^\tau=\{(X_i,A_i,Y_i)\}_{i=1}^n$ consisting of i.i.d.\ observations, we apply the estimator in \eqref{eq:pfn-mean-estimator} separately to the treated and control groups, obtaining $\widehat\mu_1(x)$ and $\widehat\mu_0(x)$, respectively. The T-learner \citep{kunzel2019metalearners} then takes their difference:
\begin{equation}
    \widehat\tau(x)
    =
    \widehat\mu_1(x)-\widehat\mu_0(x).
    \label{eq:pfn-cate-estimator}
\end{equation}

\begin{figure}[htb!]
    \centering
    \includegraphics[width=\linewidth]{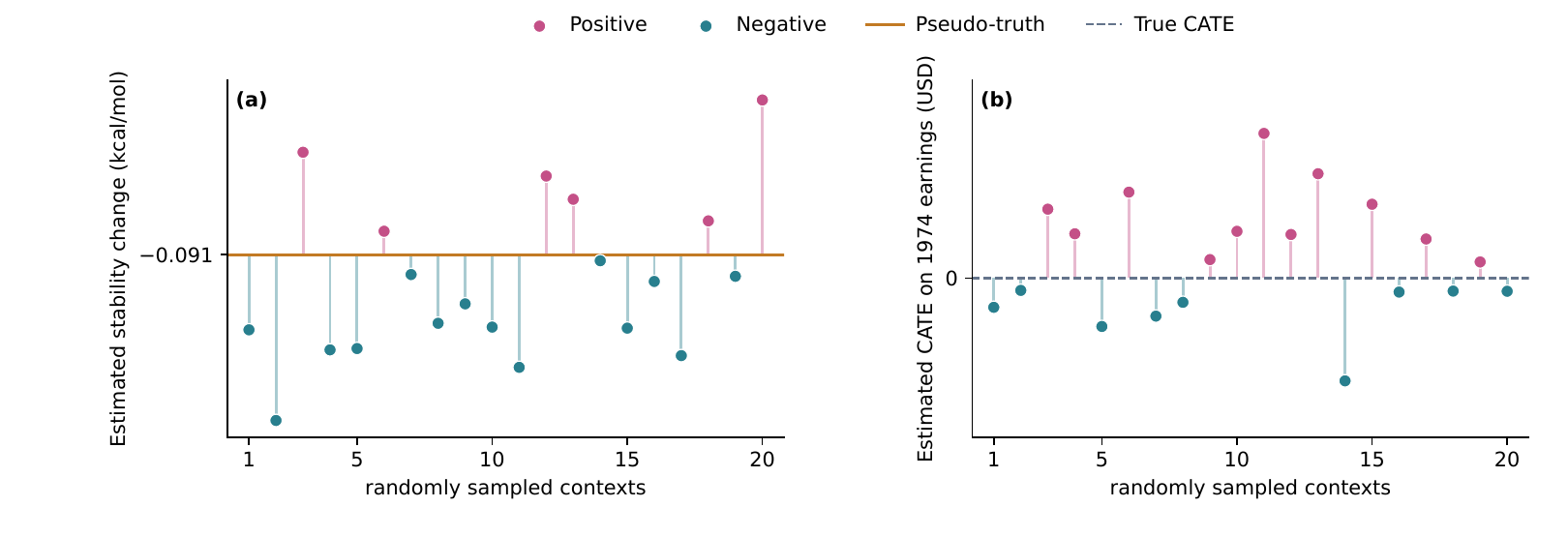}
    \caption{\textbf{Context uncertainty in two real-world examples.} (a) Estimated conditional-mean stability change for the N139V mutation in PDB 1G3P. (b) Estimated CATE of job-training assignment on pretreatment earnings in the National Supported Work experiment, where the true effect is zero. 
    Pink and teal indicate positive and negative estimates the solid orange line in (a) indicates the pseudo-truth from a LightGBM reference model fitted to the full protein data observations, approximately $-0.091$ kcal/mol. The gray dashed line in (b) marks the known zero CATE.}
    \label{fig:small-context-examples}
\end{figure}

Although tabular foundation model achieves strong prediction performance,
the point estimates produced by the tabular foundation model alone are insufficient to guide scientific decisions. We illustrate this in Figure \ref{fig:small-context-examples} with two real-data examples. 
Our first example concerns mutation-induced changes in protein stability. The feature vector $X$ is the difference between the embeddings of the wild-type and mutant protein sequences and the response is the difference in Gibbs free energy of unfolding between the mutant and wild-type proteins, measured in kcal/mol. A positive value indicates that the mutant protein is more stable than the wild type, whereas a negative value indicates that it is less stable. Here, $\mu(x)$ represents the expected stability change conditional on the mutation's feature representation. Using the protein stability dataset accompanying GeoStab \citep{xu2024improving}, we consider the N139V mutation in protein 1G3P. 

We randomly draw 20 contexts, each containing 100 labeled observations sampled without replacement from a total of 8,754 mutations. For each context, we obtain a TabPFN 2.6 estimate at the same query. A model fitted to the much larger full dataset provides an empirical reference estimate of approximately $-0.091$ kcal/mol. As shown in \Cref{fig:small-context-examples}(a), six of the 20 TabPFN estimates are positive, while the remaining 14 are negative. Thus, a decision based solely on the estimated direction of the stability change may depend on the particular observations included in the context. This sensitivity to context sampling also matters when estimates are used to assess the effects of an intervention. 

Our second example concerns a job-training programme and uses a pretreatment outcome to provide a benchmark with a known null effect. We use the National Supported Work experiment studied by \citet{lalonde1986evaluating}, with the sample compiled by \citet{dehejia1999causal}. Treatment is assignment to the training programme, and we take earnings in 1974, before treatment, as the outcome. Since later assignment cannot change income already earned, the true CATE is zero. Yet, for a fixed baseline covariate profile, the CATE estimates range from approximately $-676$ to $952$ US dollars across the 20 contexts shown in \Cref{fig:small-context-examples}(b). Interpreting their signs alone could suggest either a benefit or a harm where neither is possible.

Together, these examples show that point estimates can lead to different scientific conclusions as the context changes. Variation across contexts demonstrates context uncertainty, while the apparent displacement of estimates from their reference values raises the possibility of limited-context bias. These observations motivate uncertainty quantification that accounts for both sampling variability and potential bias before the estimated magnitude or sign is used to guide a scientific decision.

\subsection{Challenges of Uncertainty Quantification for Tabular Foundation Models}

To support scientific conclusions about conditional means and CATEs, we construct confidence intervals by bounding the estimation error. We focus on the conditional mean to illustrate the main idea and the same construction applies to CATE. A confidence interval provides a range of plausible values for the average outcome at $x$ in regression. For a fixed query $x$ and significance level $\alpha\in(0,1)$, we seek a data-dependent bound $R_\alpha(D_n;x)$ satisfying
\[
    \mathbb P\!\left\{
        |\muhat(x)-\mu(x)|\leq R_\alpha(D_n;x)
    \right\}
    \geq 1-\alpha.
\]
Using this bound as the radius gives the $1-\alpha$ confidence interval
\[
    C_\alpha(D_n;x)
    =
    \left[
        \muhat(x)-R_\alpha(D_n;x),~
        \muhat(x)+R_\alpha(D_n;x)
    \right].
\]
Its width indicates estimation precision, and an interval entirely above or below a scientifically relevant benchmark supports the corresponding conclusion about $\mu(x)$. One of the most commonly used confidence interval is the Wald interval \citep{wasserman2004all}, which uses an estimated standard error multiplied by a normal critical value as its radius:
\begin{equation*}
    C_{\mathrm{Wald}}(D_n;x)
    =
    \left[
        \muhat(x)-z_{1-\alpha/2}\widehat s(x),~
        \muhat(x)+z_{1-\alpha/2}\widehat s(x)
    \right],
    \label{eq:wald}
\end{equation*}
where $z_{1-\alpha/2}$ is the $(1-\alpha/2)$ quantile of the standard normal distribution. Here, $\widehat s(x)$ estimates the standard deviation of $\muhat(x)$ across repeated contexts and thus quantifies context variation. Approximate nominal coverage requires an accurate standard error estimate, an appropriate normal approximation, and limited-context bias that is negligible relative to the standard error.

\begin{figure}[!htb]
    \centering
    \includegraphics[width=\linewidth]{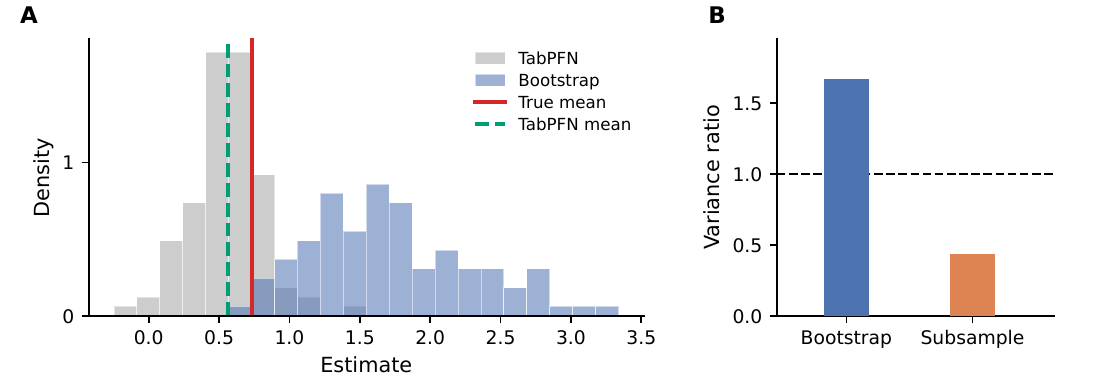}
    \caption{Context variation and limited-context bias of TabPFN 2.6 in the simulation setting with $Y=\sin(\pi X)+\varepsilon$, where independent $X,\varepsilon\sim N(0,1)$ and the context size $n=300$. (A) At $x=-1.737$, TabPFN estimates from 100 independent contexts (gray) and 100 bootstrap estimates conditional on one context (blue). (B) Ratios of mean resampling variance to the variance of TabPFN estimates across independent contexts, averaged over 20 queries in $[-3,3]$. We use the same 100 contexts to estimate both variance, with 100 resamples per method and context. Subsampling draws $m=30$ without replacement and scales variance by $m/(n-m)$.}
    \label{fig:sine-resampling-variance}
\end{figure}

Bootstrap and subsampling are often used to estimate the standard error. Ordinary bootstrap draws $n$ observations with replacement from the observed context, whereas subsampling draws $m<n$ observations without replacement \citep{efron1979bootstrap,politis1999subsampling}. We supply each resampled dataset to the foundation model as a new context and use the variation across the resulting point estimates to estimate the standard error $\widehat s(x)$ in \cref{eq:wald}. As an example, we consider TabPFN in the {\em sine} simulation setting (\cref{fig:sine-resampling-variance}). Panel~A illustrates both context variation and limited-context bias at the fixed query $x=-1.737$. The gray histogram shows TabPFN estimates across 100 independent contexts. Their empirical mean is $0.566$, compared with the true conditional mean of $0.736$, indicating a limit-context bias of $0.17$. The blue histogram shows 100 ordinary-bootstrap estimates from a single context and has a wider spread than the gray histogram. Panel~B shows that ordinary bootstrap overestimates variance on average across the query grid, whereas subsampling underestimates it. At each of 20 equally spaced queries in $[-3,3]$, we average the variance estimates from each method over the same 100 contexts and divide by the empirical variance of TabPFN estimates across those contexts. The two bars show these ratios averaged over queries: $1.64$ for ordinary bootstrap and $0.45$ for subsampling.

We next discuss possible reasons why ordinary bootstrap and subsampling misestimate context variation in our experiments. Ordinary bootstrap reuses the observed data, with some observations repeated and others omitted. TabPFN may respond differently to these repetitions than to new observations, so the bootstrap variation may not match the variation we want to estimate. Subsampling uses smaller contexts. We rescale its standard-error estimates to account for the smaller size, but the adjusted estimates still underestimate context variation in our experiments. For Wald intervals, overestimating the standard error produces wider intervals, whereas underestimating it can lead to insufficient coverage even when the point estimator's bias is small.

Accurately estimating context variation does not, however, address limited-context bias. When this bias is large relative to the standard error, a Wald interval may fail to attain nominal coverage even if the standard error is accurately estimated. Since the population target is unknown, the bias cannot be determined directly from the observed labeled context. In the next section, we therefore seek an interval construction that accounts for both context variation and limited-context bias.

\section{Method}
\label{sec:method}

TabCUQ constructs confidence intervals by combining an estimate of context variation with an empirical bias cushion for limited-context bias. We design the particular perturbation or resampling method in the following Section \ref{sec:method-design}. We develop the confidence interval construction for a conditional mean in Section \ref{sec:method-interval}, then extend it to a conditional average treatment effect in Section \ref{sec:method-cate}. 

\subsection{Perturbation design}
\label{sec:method-design}

We first consider the conditional mean at a fixed query $x$, given the observed context $D_n={(X_i,Y_i)}_{i=1}^n$. We use the wild bootstrap to perturb the outcomes while keeping all observed feature rows unchanged, where the outcome perturbation is formed by adding a randomly weighted residual to an estimated conditional mean. To avoid overfitting, we use five-fold cross-fitting to obtain the out-of-fold predictions used to construct both the residuals and the perturbed responses.

We partition the observation indices into five disjoint folds $I_1,\ldots,I_5$ of approximately equal size. For each $k=1,\ldots,5$, the other four folds form the context $D_{-k}=\{(X_j,Y_j)\}_{j\notin I_k}$. Let $\widehat{\mu}_{-k}(x)$ denote the estimator in \eqref{eq:pfn-mean-estimator} evaluated using $D_{-k}$ as the context. For each observation $i\in I_k$, we use $X_i$ as the query to obtain the out-of-fold prediction.  
Repeating this calculation for $k=1,\ldots,5$ provides exactly one out-of-fold prediction for each observation, that is, we define
\begin{equation}
\widehat{\mu}(X_i)=\widehat{\mu}_{-k}(X_i),\quad \text{for}\quad i\in I_k \quad \text{with}\quad k=1,\ldots,5.
\end{equation}
We then compute the out-of-fold residuals and subtract their sample mean from each residual before using them to generate the response perturbations. For $i\in I_k$, with $k=1,\ldots,5$, define the out-of-fold residual and the centered residual used below by
\begin{equation*}
    r_i=Y_i-\widehat{\mu}(X_i) \quad \text{and}
    \quad
    e_i=r_i-\overline r, \quad \text{with}\quad \overline r=\frac1n\sum_{j=1}^n r_j.
    \label{eq:residuals}
\end{equation*}

Then, we form each perturbed response by multiplying the centered residual $e_i$ by a random sign and adding the result to the corresponding out-of-fold prediction $\widehat{\mu}(X_i)$, for $i\in I_k$. The out-of-fold predictions and residuals are computed once and reused in all resamples. For each resample $b=1,\ldots,B$, draw an independent Rademacher multiplier $v_{i,b}$ for every observation:
\begin{equation*}
    \mathbb P(v_{i,b}=1)=\mathbb P(v_{i,b}=-1)=\frac12,
    \qquad
    1\leq i\leq n,\quad 1\leq b\leq B.
    \label{eq:method-multipliers}
\end{equation*}
Throughout the paper, we mainly use Rademacher multipliers. We may also adopt other multiplier distributions used in the wild-bootstrap literature, including the standard normal distribution and Mammen's two-point distribution, both of which have mean zero and unit variance \citep{mammen1993bootstrap,kline2012score}. For each resample $b=1,\ldots,B$, define the perturbed outcome as
\begin{equation}
    Y_{i,b}
    =
    \widehat{\mu}(X_i)+v_{i,b}e_i,
    \qquad i\in I_k,\quad k=1,\ldots,5.
    \label{eq:wildresponse}
\end{equation}
The resulting context $D_b=\{(X_i,Y_{i,b})\}_{i=1}^n$ contains the original $n$ feature vectors in the same row order, with only the outcome variables perturbed.


\subsection{Standard error, empirical bias cushion, and confidence interval}
\label{sec:method-interval}
In this section, we leverage the perturbed data set $D_b=\{(X_i,Y_{i,b})\}_{i=1}^n$ with $1\leq b\leq B$ to quantify the context variation and limited-context bias of the tabular foundation model. 
Let $\widehat{\mu}_b(x)$ denote the estimator in \eqref{eq:pfn-mean-estimator} evaluated at the fixed query $x$ using $D_b=\{(X_i,Y_{i,b})\}_{i=1}^n$ as the context, where the variation across $\{\widehat{\mu}_b(x)\}_{1\leq b\leq B}$ comes from the multiplier vectors. 

In classical regression settings, wild-bootstrap methods can consistently approximate sampling distributions under suitable regularity conditions \citep{wu1986jackknife,mammen1993bootstrap}. Motivated by these results, we use the sample standard deviation of the PFN mean estimates obtained from the perturbed contexts as a bootstrap standard-error estimate, that is, 
\begin{equation}
    \widehat s(x)
    =\left\{
        \frac1{B-1}\sum_{b=1}^B
        \bigl(\widehat\mu_b(x)-\overline\mu_B(x)\bigr)^2
    \right\}^{1/2} \quad \text{with}\quad \overline\mu_B(x)=\frac1B\sum_{b=1}^B\widehat\mu_b(x).
    \label{eq:wildse}
\end{equation}
The estimated scale $\widehat s(x)$ describes dispersion across perturbed contexts, but does not account for possible limited-context bias relative to $\mu(x)$. 

 
 To adjust for this limited-context bias, we introduce an \emph{empirical bias cushion} constructed from the same perturbed estimates:
\begin{equation}
    \widehat r(x)
    = \max_{1\leq b\leq B}
    \left|\widehat\mu_b(x)-\widehat\mu(x)\right|,
    \label{eq:radius}
\end{equation}
where $\widehat r(x)$ is the largest absolute difference between a perturbed estimate and the original estimate. We use this quantity as an empirical proxy of an upper bound for the limited-context bias. We provide theoretical support for this interpretation in a kernel-regression setting in \cref{sec:theory}. Specifically, under the stated assumptions, $\widehat r(x)$ is a high probability upper bound for the bias of the kernel regression. Computed from the observed context and the generated outcome perturbations, this cushion provides an additional margin for possible systematic error. Adding it to the margin for context variation yields a confidence interval designed to accommodate both context variation and limited-context bias.


With the standard-error estimate and the empirical bias cushion defined, we now combine them to construct an interval for the conditional mean $\mu(x)$. For a significance level $\alpha\in(0,1)$, define the TabCUQ interval by 
\begin{equation}
    C_\alpha(D_n;x)
    =\left[
    \widehat\mu(x)-R_\alpha(D_n;x),
    \widehat\mu(x)+R_\alpha(D_n;x)
    \right],
    \label{eq:finalci}
\end{equation}
with $R_\alpha(D_n;x)=z_{1-\alpha/2}\widehat s(x)+\widehat r(x)$. The term $z_{1-\alpha/2}\widehat s(x)$ addresses context uncertainty through a normal approximation based on the wild-bootstrap standard-error estimate $\widehat s(x)$. The empirical bias cushion $\widehat r(x)$ further widens the interval on both sides to account for possible limited-context bias. 
Section~4 develops the corresponding coverage analysis under a tractable kernel regression regime. Algorithm~\ref{alg:tabcuq-mean} summarizes the procedure for constructing a conditional-mean interval.


\begin{algorithm}[!htb]
    \caption{TabCUQ for a conditional mean}
    \label{alg:tabcuq-mean}
    \begingroup
    \setlength{\parindent}{0pt}
    \textbf{Input:} Context $D_n=\{(X_i,Y_i)\}_{i=1}^n$ ($n\geq5$), fixed query $x$, pretrained TFM with fixed weights $\theta$, $B\geq2$ resamples, and level $\alpha\in(0,1)$.\par
    \smallskip
    \textbf{Output:} Original estimate $\widehat\mu(x)$ and interval $C_\alpha(D_n;x)$.\par
    \begin{enumerate}[label=\textbf{Step~\arabic*:}, ref=\arabic*, leftmargin=4.1em, labelsep=0.45em, itemsep=0.45em, topsep=0.5em, parsep=0pt]
        \item Compute the original estimate $\widehat\mu(x)$ from $D_n$ using \eqref{eq:pfn-mean-estimator}.

        \item Construct the out-of-fold predictions and centered residuals once:
        \begin{enumerate}[label=(\arabic{enumi}\alph*), ref=\arabic{enumi}\alph*, leftmargin=2.7em, labelsep=0.45em, itemsep=0.2em, topsep=0.2em, parsep=0pt]
            \item Partition $\{1,\ldots,n\}$ into five nearly equally sized disjoint folds $I_1,\ldots,I_5$.

            \item For each $k=1,\ldots,5$, use $D_{-k}=\{(X_j,Y_j):j\notin I_k\}$ as the context to compute $\widehat\mu_{-k}(X_i)$ and set $\widehat\mu(X_i)\gets\widehat\mu_{-k}(X_i)$ for $i\in I_k$.

            \item Compute and center the residuals for $i=1,\ldots,n$:
            \[
                r_i\gets Y_i-\widehat\mu(X_i),
                \qquad \overline r\gets\frac1n\sum_{j=1}^n r_j,
                \qquad e_i\gets r_i-\overline r.
            \]
        \end{enumerate}

        \item For each resample $b=1,\ldots,B$:
        \begin{enumerate}[label=(\arabic{enumi}\alph*), ref=\arabic{enumi}\alph*, leftmargin=2.7em, labelsep=0.45em, itemsep=0.2em, topsep=0.2em, parsep=0pt]
            \item Draw $v_{i,b}\in\{-1,+1\}$ with equal probabilities, independently across $i,b$ and of the data and fold partition.

            \item Set $Y_{i,b}\gets\widehat\mu(X_i)+v_{i,b}e_i$ for $i=1,\ldots,n$.

            \item Form $D_b\gets\{(X_i,Y_{i,b})\}_{i=1}^n$ in the original row order and compute $\widehat\mu_b(x)$ from $D_b$.
        \end{enumerate}

        \item Compute the standard-error estimate:
        \[
            \overline\mu_B(x)\gets\frac1B\sum_{b=1}^B\widehat\mu_b(x),
            \qquad \widehat s(x)\gets\left\{\frac1{B-1}\sum_{b=1}^B\bigl(\widehat\mu_b(x)-\overline\mu_B(x)\bigr)^2\right\}^{1/2}.
        \]

        \item Compute the empirical bias cushion $\widehat r(x)\gets\max_{1\leq b\leq B}|\widehat\mu_b(x)-\widehat\mu(x)|$.

        \item Form the interval and return $\widehat\mu(x)$ and $C_\alpha(D_n;x)$:
        \[
            \begin{aligned}
                R_\alpha(D_n;x)&\gets z_{1-\alpha/2}\widehat s(x)+\widehat r(x),\\
                C_\alpha(D_n;x)&\gets[\widehat\mu(x)-R_\alpha(D_n;x),\,\widehat\mu(x)+R_\alpha(D_n;x)].
            \end{aligned}
        \]
    \end{enumerate}
    \endgroup
\end{algorithm}

\subsection{Inference for conditional average treatment effects}
\label{sec:method-cate}

In the following section, we generalize our proposal to uncertainty quantification for CATE. Recall the CATE estimation using the T-learner in Section~2 \citep{kunzel2019metalearners}, which estimates the conditional mean in each treatment arm and takes the difference between the two estimates. Let $D_n^\tau=\{(X_i,A_i,Y_i)\}_{i=1}^n$ be a context consisting of i.i.d.\ observations, where $X_i$ denotes pretreatment covariates, $A_i\in\{0,1\}$ indicates treatment, and $Y_i$ is the observed outcome. Under consistency, unconfoundedness, and overlap as stated in Section~2, the CATE is $\tau(x)=\mathbb E[Y(1)-Y(0)\mid X=x]=\mu_1(x)-\mu_0(x)$, where $\mu_a(x)=\mathbb E[Y\mid X=x,A=a]$ denotes the conditional mean for the treatment assignment $a$. For each arm $a\in\{0,1\}$, we apply the pretrained model with $D_a=\{(X_i,Y_i):A_i=a\}$ as the context to obtain the conditional-mean estimate $\widehat\mu_a(x)$ in \eqref{eq:pfn-mean-estimator}. The original T-learner estimate is $\widehat\tau(x)=\widehat\mu_1(x)-\widehat\mu_0(x)$.

We apply Steps~2--3 of Algorithm~\ref{alg:tabcuq-mean} separately to $D_0$ and $D_1$. Cross-fitting and residual centering are performed within each arm. Let $\widehat\mu_{a,b}(x)$ denote the resulting prediction from arm $a$ in resample $b$. The perturbed CATE estimate is
\begin{equation}
    \widehat\tau_b(x)
    =\widehat\mu_{1,b}(x)-\widehat\mu_{0,b}(x),
    \qquad b=1,\ldots,B.
    \label{eq:method-cate-replicates}
\end{equation}
Then following \eqref{eq:wildse} and \eqref{eq:radius}, we compute $\widehat s_\tau(x)$ as the sample standard deviation of the perturbed CATE estimates and $\widehat r_\tau(x)$ as their maximum absolute deviation from $\widehat\tau(x)$. The CATE interval is
\begin{equation}
    C_{\alpha,\tau}(D_n^\tau;x)
    =\left[
        \widehat\tau(x)-
        \left\{z_{1-\alpha/2}\widehat s_\tau(x)+\widehat r_\tau(x)\right\},\widehat\tau(x)+
        \left\{z_{1-\alpha/2}\widehat s_\tau(x)+\widehat r_\tau(x)\right\}
    \right].
    \label{eq:method-cate-interval}
\end{equation}
Algorithm~\ref{alg:tabcuq-cate} summarizes the procedure for constructing a CATE interval.

\begin{algorithm}[!htb]
    \caption{TabCUQ for a CATE using a T-learner}
    \label{alg:tabcuq-cate}
    \begingroup
    \setlength{\parindent}{0pt}
    \textbf{Input:} Context $D_n^\tau=\{(X_i,A_i,Y_i)\}_{i=1}^n$ with at least five observations in each arm, fixed query $x$, pretrained TFM with fixed weights $\theta$, $B\geq2$ resamples, and level $\alpha\in(0,1)$.\par
    \smallskip
    \textbf{Output:} Original CATE estimate $\widehat\tau(x)$ and interval $C_{\alpha,\tau}(D_n^\tau;x)$.\par
    \begin{enumerate}[label=\textbf{Step~\arabic*:}, ref=\arabic*, leftmargin=4.1em, labelsep=0.45em, itemsep=0.45em, topsep=0.5em, parsep=0pt]
        \item For each $a\in\{0,1\}$, form $D_a=\{(X_i,Y_i):A_i=a\}$ and apply Steps~1--3 of Algorithm~\ref{alg:tabcuq-mean} to obtain $\widehat\mu_a(x)$ and $\{\widehat\mu_{a,b}(x)\}_{b=1}^B$, using independent multipliers across arms.

        \item Compute the original and perturbed CATE estimates:
        \[
            \begin{aligned}
                \widehat\tau(x)
                &\gets\widehat\mu_1(x)-\widehat\mu_0(x),\\
                \widehat\tau_b(x)
                &\gets\widehat\mu_{1,b}(x)-\widehat\mu_{0,b}(x),
                \qquad b=1,\ldots,B.
            \end{aligned}
        \]

        \item Compute the standard-error estimate $\widehat s_\tau(x)$ and empirical bias cushion $\widehat r_\tau(x)$ from $\{\widehat\tau_b(x)\}_{b=1}^B$ using \eqref{eq:wildse} and \eqref{eq:radius}, respectively, with $\widehat\tau(x)$ as the original estimate. Form the interval in \eqref{eq:method-cate-interval} and return $\widehat\tau(x)$ and $C_{\alpha,\tau}(D_n^\tau;x)$.
    \end{enumerate}
    \endgroup
\end{algorithm}

For average treatment effects, semiparametric methods based on influence functions allow valid inference with flexible estimates of nuisance functions under suitable conditions \citep{hines2022demystifying}. With continuously distributed covariates, however, a conditional mean or CATE at a fixed query is generally not pathwise differentiable in an unrestricted nonparametric model. The usual regular root-$n$ theory for average treatment effects therefore does not apply directly. Although orthogonality can reduce sensitivity to nuisance-estimation errors, it does not by itself eliminate the bias introduced by estimating the conditional effect. For the tabular foundation models considered here, reliable inference must account for both \emph{context variation}, arising from the randomness of the finite labeled sample supplied to the model, and \emph{limited-context bias}, the systematic difference between the estimator's expectation across sampled contexts and the target.

In our experiments, we use a T-learner: the pretrained model estimates the conditional mean outcome separately under treatment and control, and their difference estimates CATE. Other learners could also be considered. For example, the DR-learner combines estimated outcome regressions and treatment probabilities to construct pseudo-outcomes, which are then regressed on the covariates to estimate CATE \citep{kennedy2023towards}. Such an approach could reduce the contribution of nuisance-estimation errors, although the final regression step may still introduce approximation or regularization bias. Here, we focus on uncertainty quantification for the T-learner, using TabCUQ to estimate context variation and provide an empirical bias cushion. Extending this construction to other learners would require accounting for their additional estimation steps and assessing the resulting interval coverage.

\section{Theory}
\label{sec:theory}

To clarify the statistical reasoning behind TabCUQ, we study outcome perturbations in the setting of kernel regression. This tractable setting allows us to examine how the perturbed estimates relate to sampling variation and how the bias cushion provides an upper bound for the nonparametric bias. The purpose of our theoretical analysis is to explain the rationale for TabCUQ's interval construction, especially the role of the empirical bias cushion, rather than to establish theoretical guarantees for pretrained tabular foundation models.

We first introduce the basic setup of kernel regression and then adapt the perturbation procedure described in Sections~\ref{sec:method-design} and~\ref{sec:method-interval} to this setting. Let $\{(X_i,Y_i)\}_{i=1}^n$ be i.i.d. observations of $(X,Y)$, where $X\in\mathbb R^d$, $Y\in\mathbb R$, and $\mathbb E|Y|<\infty$. Throughout this section, we fix our query feature at $x\in\mathbb R^d$ and our target is the regression function $\mu(x)=\mathbb E[Y\mid X=x]$. Defining $\varepsilon_i=Y_i-\mu(X_i)$ gives the representation
\[
    Y_i=\mu(X_i)+\varepsilon_i,
    \qquad
    \mathbb E[\varepsilon_i\mid X_i]=0.
\]
We write $\sigma^2(x)=\mathbb E[\varepsilon_i^2\mid X_i=x]$ and assume $\sigma^2(x)$ is continuous at $x$.

For a nonnegative kernel $K$ and bandwidth $h_n>0$, define the kernel weights
\begin{equation}
    w_i(x)
    =
    \frac{K((X_i-x)/h_n)}
         {\sum_{j=1}^n K((X_j-x)/h_n)},
    \qquad 1\leq i\leq n,
    \label{eq:tcq-kernel-weights}
\end{equation}
with all weights set to zero when the denominator is zero. When the denominator is positive, the Nadaraya-Watson estimator is the solution to the local-constant least-squares problem
\begin{equation}
    \muhat(x)
    =
    \operatorname*{arg\,min}_{a\in\mathbb R}
    \sum_{i=1}^n w_i(x)(Y_i-a)^2
    =
    \sum_{i=1}^n w_i(x)Y_i.
    \label{eq:tcq-estimator-scale}
\end{equation}
Intuitively, equation~\eqref{eq:tcq-estimator-scale} estimates $\mu(x)$ by a weighted average of the responses, with weights determined by the kernel, the bandwidth, and the feature locations relative to $x$.

In this kernel-regression setting, we construct the outcome perturbations without cross-fitting. We form the residuals $r_i=Y_i-\widehat\mu(X_i)$, where $\widehat\mu(X_i)$ is the estimator in \eqref{eq:tcq-estimator-scale} evaluated at $X_i$, and center them as $e_i=r_i-\bar r$, where $\bar r=n^{-1}\sum_{j=1}^n r_j$. These fitted values and residuals are computed once and reused in all perturbations. For each perturbation $b=1,\ldots,B$, we generate $Y_{i,b}=\widehat\mu(X_i)+v_{i,b}e_i$ using the Rademacher multipliers described in Section~\ref{sec:method-design}. Keeping the features and kernel weights fixed, we replace the responses in \eqref{eq:tcq-estimator-scale} with $Y_{i,b}$ to obtain $\widehat\mu_b(x)=\sum_{i=1}^n w_i(x)Y_{i,b}$. As in Section~\ref{sec:method-interval}, we compute the standard-error estimate $\widehat s(x)$ from the perturbed estimates $\{\widehat\mu_b(x)\}_{1\leq b\leq B}$. The empirical bias cushion $\widehat r(x)$ is the largest absolute difference between all perturbed estimates $\{\widehat\mu_b(x)\}_{1\leq b\leq B}$ and the original estimate $\widehat\mu(x)$. We combine these quantities to form the interval $C_\alpha(D_n;x)$ in \eqref{eq:finalci}. We analyze this construction under the following assumptions.

\begin{assumption}[Kernel regression assumptions]
    \label{ass:kr-setting}
    \begin{enumerate}
        \item The observations $(X_i,Y_i)$ are i.i.d., with $X_i\in\mathbb R^d$ and satisfy $Y_i=\mu(X_i)+\varepsilon_i$ with $\mathbb E|\mu(X)|<\infty$. The design distribution has a density $f_X$ that is continuous at the fixed query $x$, with $0<f_X(x)<\infty$. There exists a finite constant $C>0$ such that
        \[ 
            \mathbb E[\varepsilon_i\mid X_i]=0,\qquad \mathbb E[\varepsilon_i^2\mid X_i]=\sigma^2(X_i),\qquad \mathbb E[\varepsilon_i^4\mid X_i]\le C \quad\text{almost surely},
        \]
        where $\sigma^2(x)>0$ is continuous at $x$.
        \item The regression function $\mu$ is H\"older smooth in the following sense:: For some $\beta>0$ and finite $L>0$, the regression function $\mu:\mathbb R^d\to\mathbb R$ is $\ell$ times continuously differentiable with $\ell=\lceil\beta\rceil-1$, and satisfies
        \[ 
            |D^\gamma\mu(u)-D^\gamma\mu(v)|\le L\|u-v\|^{\beta-\ell},\qquad u,v\in\mathbb R^d,\quad |\gamma|=\ell. 
        \]
where $\gamma=(\gamma_1,\ldots,\gamma_d)\in\mathbb N_0^d$ with $\mathbb N_0^d$ denoting the set of $d$-tuples of nonnegative integers, $|\gamma|=\sum_{j=1}^d\gamma_j$, $D^\gamma$ denotes the corresponding partial derivative with $D^{\mathbf 0}\mu=\mu$ and $\|\cdot\|$ is the Euclidean norm. 
        \item There exist constants $0<r_0\le1$ and $0<K_{\min}\le K_{\max}<\infty$ such that
        \[ 
            K_{\min}\mathbf 1\{\|u\|\le r_0\}\le K(u)\le K_{\max}\mathbf 1\{\|u\|\le1\},\qquad u\in\mathbb R^d. 
        \]
        The kernel $K$ is of order $\ell=\lceil\beta\rceil-1$, that is,  for every multi-index $\gamma=(\gamma_1,\ldots,\gamma_d)\in\mathbb N_0^d$ with $|\gamma|\le\ell$,
        \[ \int_{\mathbb R^d}K(u)\,du=1,\qquad \int_{\mathbb R^d}\prod_{j=1}^d u_j^{\gamma_j} K(u)\,du=0,\quad 1\le|\gamma|\le\ell. \]
In addition, the kernel satisfies $ \int_{\mathbb R^d}\|u\|^\beta|K(u)|\,du<\infty.$

        \item The deterministic bandwidth satisfies
        \[
            h_n\to0,
            \qquad
            \frac{nh_n^d}{\log n}\to\infty.
        \]
    \end{enumerate}
\end{assumption}
In the above assumption, the H\"older smoothness of the function $\mu(\cdot)$ and kernel assumptions are standard in nonparametric estimation; see \cite{tsybakov2009}.

In the following, we shall use the Nadaraya--Watson estimator
as an analytically tractable example to provide reasoning for our TabCUQ proposal. We use the terms
\emph{context variation} and \emph{limited context bias}
to connect the interpretation of its estimation error with
our discussion of tabular foundation models. For the
Nadaraya--Watson estimator, these two components admit an
explicit characterization through the exact decomposition
\begin{equation}
    \widehat\mu_{\mathrm{NW}}(x)-\mu(x)
    =
    \underbrace{
        \widehat\mu^{\mathrm{or}}(x)-\mu(x)
    }_{\text{context variation}}
    +
    \underbrace{
        b_{n}(x)
    }_{\text{limited context bias}}
\label{eq:tcq-error-decomposition}
\end{equation}
where $\widehat\mu^{\mathrm{or}}(x)$ denotes an oracle estimator of $\mu$ and $b_n(x)$ denotes the bias component, defined as
\begin{equation}
    \widehat\mu^{\mathrm{or}}(x)
    :=
    \mu(x)+\sum_{i=1}^n w_i(x)\varepsilon_i \quad \text{and}\quad b_n(x)=\sum_{i=1}^n w_i(x)\{\mu(X_i)-\mu(x)\}.
    \label{eq:tcq-oracle}
\end{equation}
Conditional on the observed covariates, context variation
corresponds to the mean-zero fluctuation induced by the outcome
noise. Limited context bias corresponds to the discrepancy
between the weighted regression values within the context
and the target $\mu(x)$. The oracle estimator removes this
bias while preserving context variation. The two quantities used by TabCUQ correspond to the two terms in \eqref{eq:tcq-error-decomposition}: the standard-error estimate $\widehat s(x)$ estimates the scale $\rho_n$ of the oracle fluctuation, while the empirical bias cushion $\widehat r(x)$ provides a high-probability upper bound for the bias $b_n(x)$.





Under Assumption~\ref{ass:kr-setting}, the oracle estimator satisfies
\begin{equation}
    \frac{\widehat\mu^{\mathrm{or}}(x)-\mu(x)}{\rho_n}
    \xrightarrow{d}
    N(0,1)
    \qquad \text{with}\qquad 
        \rho_n
    =
    \sigma(x)\sqrt{\frac{\nu_K}{nh_n^d f_X(x)}},
    \label{eq:tcq-sampling-scale}
\end{equation}
where $\nu_K=\int_{\mathbb R^d}K(u)^2\,du$ and  $\rho_n$ describes the asymptotic sampling scale of $\widehat\mu^{\mathrm{or}}(x)-\mu(x)$.

\begin{theorem}[An upper bound on bias]
    \label{thm:tcq-bias}
    Under Assumption~\ref{ass:kr-setting}, suppose the number $B$ of perturbations satisfy 
    \begin{equation}
        \frac{nh_n^{2\beta+d}}{\log B}\longrightarrow0.
    \label{eq: Bn requirement}
    \end{equation}
    Then, as $n\to\infty$, the empirical bias cushion $\widehat r(x)$ defined in \eqref{eq:radius} satisfies
    \begin{equation}
        \Pp\!\left\{
            |b_n(x)|\le\widehat r(x)
        \right\}
        \longrightarrow1.
        \label{eq:tcq-bias-bound}
    \end{equation}
\end{theorem}

Theorem~\ref{thm:tcq-bias} shows that the empirical bias cushion is a high probability upper bound for the bias component. On the event $|b_n(x)|\le\widehat r(x)$, the decomposition in \eqref{eq:tcq-error-decomposition} gives $|\muhat(x)-\mu(x)|\le|\muhat^{\mathrm{or}}(x)-\mu(x)|+\widehat r(x)$. Together with the limiting distribution in \eqref{eq:tcq-sampling-scale}, the bias bound therefore yields a pointwise coverage guarantee for $C_\alpha(D_n;x)$. The following theorem states this guarantee and bounds the interval width.



\begin{theorem}[Pointwise coverage and interval width]
    \label{thm:kr-pointwise-coverage}
    Under Assumption~\ref{ass:kr-setting} and suppose that $B$ satisfies \eqref{eq: Bn requirement}.
    For fixed $\alpha\in(0,1)$, the confidence interval $C_\alpha(D_n;x)$ satisfies
    \begin{equation}
        \liminf_{n\to\infty}
        \mathbb P\bigl\{\mu(x)\in C_\alpha(D_n;x)\bigr\}
        \ge1-\alpha.
        \label{eq:kr-pointwise-coverage}
    \end{equation}
    Moreover,
    \begin{equation}
        \operatorname{length}\{C_\alpha(D_n;x)\}
        =
        O_p\!\left(\rho_n\sqrt{\log B}\right).
        \label{eq:kr-interval-width}
    \end{equation}
\end{theorem}

Theorem~\ref{thm:kr-pointwise-coverage} establishes asymptotic pointwise coverage of at least $1-\alpha$ and provides an upper bound on the interval width. The usual kernel-regression error scale combines sampling variability and bias, $(nh_n^d)^{-1/2}+h_n^\beta$. The width bound depends on this budget through the factor $\sqrt{\log B_n}$, highlighting a tradeoff among bias accommodation, interval width, and computational cost. The bias component depends on the unknown regression function and is generally harder to estimate than the standard error. Balancing these terms gives $h_n\asymp n^{-1/(2\beta+d)}$ and the classical optimal pointwise length benchmark $n^{-\beta/(2\beta+d)}$. Under the theorem's conditions, choosing this bandwidth and $B=\lceil n^c\rceil$ for any fixed $c>0$ yields $\operatorname{length}\{C_\alpha(D_n;x)\}=O_p(n^{-\beta/(2\beta+d)}\sqrt{\log n})$. Our length bound is therefore near-optimal up to a logarithmic factor, which arises from taking the maximum over perturbations to accommodate bias without directly estimating it.

\section{Simulation Studies}
\label{sec:experiments}

\subsection{Experimental design and benchmarks}
\label{sec:simulation-setup}

We evaluate the coverage and width of nominal $95\%$ confidence intervals for conditional means $\mu(x)$ and conditional average treatment effects $\tau(x)$ in simulations where the target values are known. The primary experiments use TabPFN 2.6 with eight ensemble members. Each Monte Carlo repetition generates an independent labeled context, including both features and outcomes, while the queries, pretrained model, and model seed remain fixed. Table~\ref{tab:simulation-dgps} summarizes the main data-generating processes. For conditional means, we consider univariate and multivariate regression models with different mean functions, outcome noise distributions, and covariate structures. For CATEs, we consider nine designs that vary the baseline response, treatment-effect function, outcome noise, and treatment assignment.

\begin{table}[htb!]
\centering
\small
\setlength{\tabcolsep}{3pt}
\renewcommand{\arraystretch}{1.15}
\caption{Synthetic data-generating processes for conditional means and CATE. U and M stand for paired univariate and multivariate designs, respectively; C stands for the CATE designs.}
\label{tab:simulation-dgps}
\begin{tabular}{@{}>{\raggedright\arraybackslash}p{.28\linewidth}>{\raggedright\arraybackslash}p{.15\linewidth}>{\raggedright\arraybackslash}p{.25\linewidth}>{\raggedright\arraybackslash}p{\dimexpr.32\linewidth-18pt\relax}@{}}
\toprule
\multicolumn{4}{@{}l}{\textit{A. Conditional means: $Y=\mu(X)+\varepsilon(X)$}}\\[2pt]
\multicolumn{4}{@{}p{\linewidth}@{}}{\footnotesize Univariate: $X\sim N(0,1)$. Multivariate: $X\sim N(0,\Sigma_\rho)$, with $p\in\{5,10\}$ and $\rho=0$ for M1--M8. M1 also uses $\rho=0.5$; M9 uses $p=2$ and $\rho\in\{0,0.5\}$.}\\[3pt]
Design & Univariate mean & Multivariate mean & Outcome noise\\
\midrule
Linear (U1 / M1) & $x$ & $\sum_{j=1}^{p}x_j$ & $Z$\\
Quadratic (U2 / M2) & $x^2$ & $\sum_{j=1}^{p}x_j^2$ & $Z$\\
Bimodal (U3 / M3) & $x$ & $\sum_{j=1}^{p}x_j$ & $1.5U+Z$\\
Heteroskedastic linear\newline (U4 / M4) & $x$ & $\sum_{j=1}^{p}x_j$ & 1D: $(1+|x|)Z$\newline Multi: $\bigl(1+\frac{|\sum_jx_j|}{p}\bigr)Z$\\[2pt]
Absolute value (U5 / M5) & $|x|$ & $\sum_{j=1}^{p}|x_j|$ & $Z$\\
Sine (U6 / M6) & $\sin(\pi x)$ & $\sum_{j=1}^{p}\sin(\pi x_j)$ & $Z$\\
Step (U7 / M7) & $\operatorname{sgn}_+(x)$ & $\sum_{j=1}^{p}\operatorname{sgn}_+(x_j)$ & $Z$\\
Heteroskedastic quadratic\newline (U8 / M8) & $x^2$ & $\sum_{j=1}^{p}x_j^2$ & 1D: $(1+|x|)Z$\newline Multi: $\bigl(1+\frac{|\sum_jx_j|}{p}\bigr)Z$\\[2pt]
Interaction (M9) & -- & $x_1+x_2+x_1x_2$ & $Z$\\
\midrule
\multicolumn{4}{@{}l}{\textit{B. CATE: $Y=\mu_0(X)+A\tau(X)+\varepsilon(X)$}}\\[2pt]
\multicolumn{4}{@{}p{\linewidth}@{}}{\footnotesize $X\sim N(0,1)$ and $A\mid X=x\sim\operatorname{Bernoulli}\{e(x)\}$. Common settings are $\varepsilon(x)=Z$ and $e(x)=0.5$; a dash in the final column denotes these settings.}\\[3pt]
Design & Baseline $\mu_0(x)$ & Effect $\tau(x)$ & Noise / assignment change\\
\midrule
C1: Linear effect & $x$ & $x$ & --\\
C2: Null effect & $x$ & $0$ & --\\
C3: Opposite slopes & $-x$ & $2x$ & --\\
C4: Quadratic baseline & $x^2$ & $x$ & --\\
C5: Bimodal noise & $x$ & $x$ & $\varepsilon(x)=1.5U+Z$\\
C6: Heteroskedastic noise & $x$ & $x$ & $\varepsilon(x)=(1+|x|)Z$\\
C7: Nonlinear effect & $x$ & $x^2$ & --\\
C8: Treatment imbalance & $x$ & $x$ & $e(x)=0.2$\\
C9: Weak overlap & $x$ & $x$ & $e(x)=\{1+\exp(x)\}^{-1}$\\
\bottomrule
\end{tabular}
\par\smallskip
\begin{minipage}{\linewidth}
\footnotesize
Here $Z\sim N(0,1)$ and $U\in\{-1,1\}$ with equal probabilities are independent of the covariates and of each other. We use $\operatorname{sgn}_+(z)=1$ for $z\geq0$ and $-1$ otherwise. The covariance $\Sigma_\rho$ has unit diagonal and off-diagonal entries $\rho$. In Panel A, a noise formula without a dimension label applies to both univariate and multivariate versions; the interaction design is multivariate only.
\end{minipage}
\end{table}

{We compare TabCUQ with confidence intervals based solely on the wild-bootstrap standard error (referred to as \emph{Wild bootstrap}). Both methods use the same point estimate and estimated standard error, so this comparison isolates the contribution of the empirical bias cushion. For CATE estimation, we use a T-learner with separate contexts for the two treatment arms. We also consider an oracle bias-aware (OBA) benchmark that uses the known target value and independent datasets generated from the true data-generating distribution to estimate the bias and sampling standard deviation of the original estimator. These quantities are used to construct benchmark intervals that provide a reference for interval width. The OBA method relies on information available in our simulations but unavailable in practice.}

At each query point, coverage is estimated by the proportion of Monte Carlo confidence intervals containing the true target, while interval width is averaged across repetitions. We also evaluate the standard-error estimates. We report both pointwise results and averages over prespecified query regions. For the univariate and CATE settings, we use 20 equally spaced points in $[-3,3]$, including eight edge points satisfying \(|x|\geq 2\). For multivariate regression, we consider 20 query points in each of three Mahalanobis probability strata (0–50\%, 50–90\%, and 90–99\%). The main univariate experiments use $S=300$ independent contexts of size $n=300$, with $B=300$ wild-bootstrap perturbations per context. For multivariate regression, we use $S=100$ independent contexts of size $n=300$, with $B=100$ wild-bootstrap perturbations per context. The CATE experiments use the same numbers of contexts and perturbations, with a total sample size of $n=600$ across the two treatment arms. Complete description of the simulation settings and evaluation procedures are given in Appendix~B.

\FloatBarrier
\graphicspath{{overleaf_figures/}{../overleaf_figures/}}

\subsection{Main results}
\label{sec:main-result}

\begin{figure}[htb!]
    \centering
    \includegraphics[width=\linewidth]{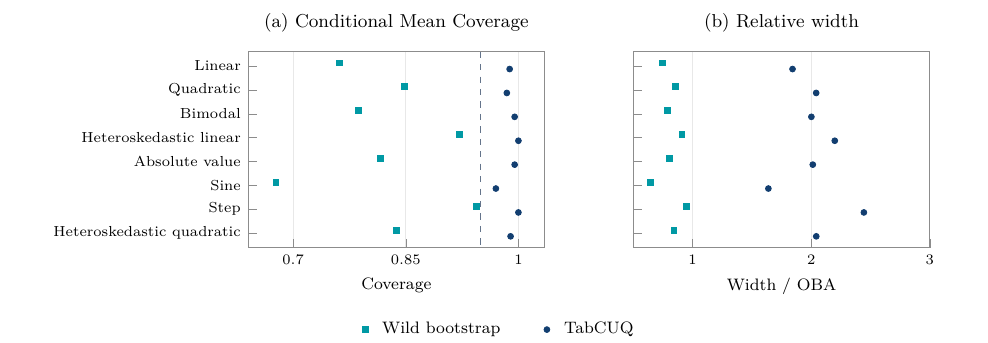}
    \caption{Univariate conditional-mean interval performance for TabPFN 2.6 across eight designs, with $S=300$, $n=300$, and $B=300$. (a) Coverage averaged over the eight edge queries satisfying $|x|\geq2$. (b) Mean pointwise interval-width/OBA-width ratio over the same queries. Dashed reference lines mark $0.95$ coverage and a width ratio of one.}
    \label{fig:mean-uq-summary}
\end{figure}

\begin{figure}[htb!]
    \centering
    \includegraphics[width=\linewidth]{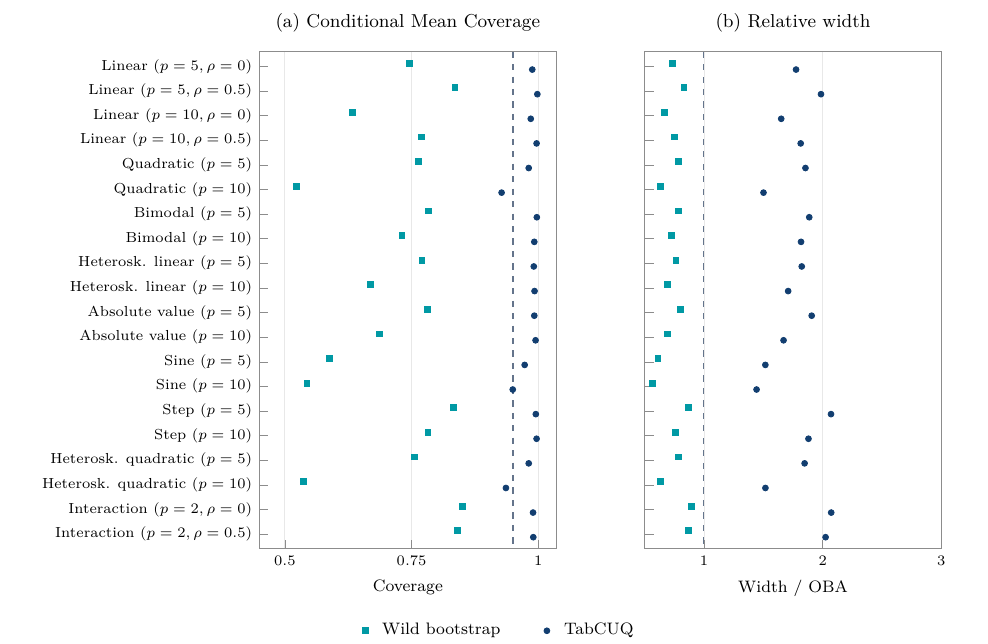}
    \caption{Multivariate conditional-mean interval performance for TabPFN 2.6 across 20 settings, with $S=100$ independent contexts of size $n=300$ and $B=100$. (a) Coverage averaged over 20 fixed queries in the $90$--$99\%$ Mahalanobis stratum. (b) Mean pointwise interval-width/OBA-width ratio over the same queries. 
    Dashed reference lines mark $0.95$ coverage and a width ratio of one.}
    \label{fig:multivariate-main}
\end{figure}

TabCUQ substantially improves conditional-mean coverage across the univariate and multivariate designs. For the eight univariate designs, coverage averaged over the edge queries ranges from $0.677$ to $0.945$ for Wild bootstrap and from $0.970$ to $1.000$ for TabCUQ; see Figure~\ref{fig:mean-uq-summary}. The empirical bias cushion thus raises the regional average coverage above the nominal $0.95$ level in all eight designs. The greatest improvement occurs in the sinusoidal design, where the coverage increases from $0.677$ to $0.970$. TabCUQ also improves coverage in all 20 multivariate settings evaluated in the $90$--$99\%$ Mahalanobis stratum; see Figure~\ref{fig:multivariate-main}. In these settings, the average coverage ranges from $0.523$ to $0.851$ for the Wild bootstrap and from $0.928$ to $0.998$ for TabCUQ. The details are given in Table~7 in Appendix~D.

\begin{figure}[htb!]
    \centering
    \includegraphics[width=\linewidth]{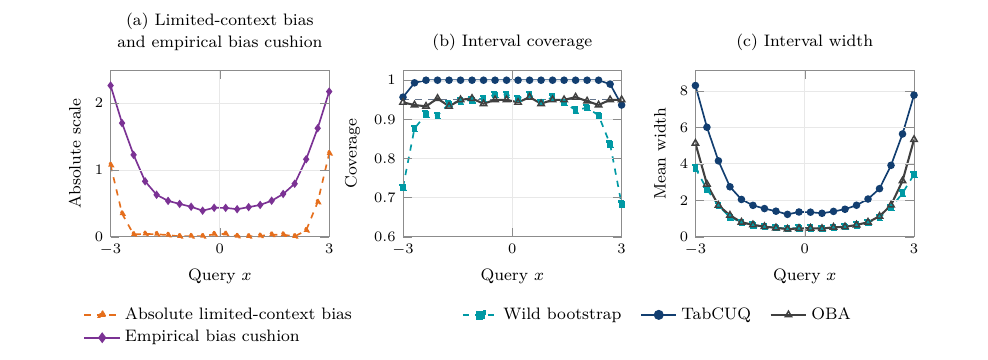}
    \caption{Pointwise conditional-mean results in the quadratic design for TabPFN 2.6, with $S=300$, $n=300$, and $B=300$. (a) Absolute Monte Carlo limited-context bias (orange dashed) and mean empirical bias cushion (blue solid). (b) Coverage of wild-bootstrap and TabCUQ intervals; the horizontal dashed line marks $0.95$. (c) Mean interval widths for wild bootstrap, TabCUQ, and OBA. The cushion in (a) is the added component of the TabCUQ half-width.}
    \label{fig:univariate}
\end{figure}

Detailed results under the quadratic design are presented in Figure~\ref{fig:univariate} to illustrate the role of the empirical bias cushion in improving coverage in queries with substantial limited-context bias. Figure~\ref{fig:univariate}(a) shows that the absolute Monte Carlo bias is small near the center of the query grid but increases sharply toward its endpoints. The mean empirical bias cushion also increases toward these endpoints, providing a larger addition to the interval half-width where the bias is most pronounced. At $x=3$, the absolute Monte Carlo bias is $1.247$, and the mean empirical bias cushion is $2.171$. The Wild bootstrap interval achieves coverage of only $0.683$ at this query. Adding the empirical bias cushion increases coverage to $0.937$, close to the nominal $0.95$ level; see Figure~\ref{fig:univariate}(b). Figure~\ref{fig:univariate}(c) shows the corresponding increase in mean interval width, with the largest absolute increases near the endpoints of the grid. TabCUQ retains the original point estimate as the interval center and improves coverage by adding the empirical bias cushion for limited-context bias to the interval half-width.

\begin{figure}[htb!]
    \centering
    \includegraphics[width=\linewidth]{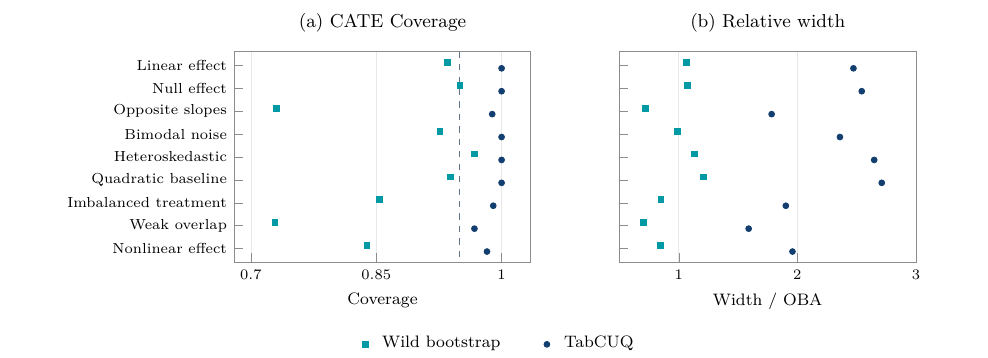}
    \caption{CATE interval performance for TabPFN 2.6 across nine designs, with $S=100$, total $n=600$, and $B=100$. (a) Coverage averaged over the eight edge queries satisfying $|x|\geq2$. (b) Mean pointwise interval-width/OBA-width ratio over the same queries. Dashed reference lines in (a,b) mark $0.95$ coverage and a width ratio of one.}
    \label{fig:cate-main}
\end{figure}

The CATE experiments reveal an additional source of variation in coverage arising from the combination of arm-specific estimation biases. Across the nine CATE designs, edge-average coverage ranges from $0.729$ to $0.968$ for Wild bootstrap intervals and from $0.968$ to $1.000$ for TabCUQ; see Figure~\ref{fig:cate-main}. Let \(b_a(x)\) denote the bias of the estimated conditional mean in treatment arm $a$. The bias of the estimated CATE then satisfies \(b_\tau(x)=b_1(x)-b_0(x)\), so the arm-specific biases may cancel or reinforce each other. In the null-effect design, the edge-average absolute CATE bias is $0.062$, and the Wild bootstrap interval achieves coverage of $0.950$. In contrast, the opposite-slopes design produces an edge-average absolute CATE bias of $0.537$, and Wild bootstrap coverage falls to $0.730$. The corresponding standard-error calibration ratios are similar, at $1.103$ and $1.107$, respectively. Adding the empirical bias cushion increases coverage in the opposite-slopes design to $0.989$. Our CATE experiments also examine nonlinear treatment effects and changes in treatment assignment. Wild bootstrap coverage falls to $0.854$ under treatment imbalance, $0.839$ for a nonlinear effect, and $0.729$ under weak overlap. TabCUQ increases coverage to $0.990$, $0.983$, and $0.968$, respectively. Weak overlap remains particularly challenging at individual queries. At $x=3$, where treated observations are sparse, coverage increases from $0.540$ to only $0.890$ after adding the empirical bias cushion.

We also compare interval widths with an oracle benchmark (OBA) constructed using the Monte Carlo bias and standard deviation. This benchmark provides a reference for interval efficiency using information unavailable from a single observed context. Across the univariate conditional-mean designs, the mean pointwise TabCUQ/OBA width ratios range from $1.64$ to $2.44$; the corresponding ratios range from $1.443$ to $2.072$ in the multivariate tail experiments. For CATE estimation, the ratios of regional mean TabCUQ width to regional mean OBA width range from $1.59$ to $2.71$ across the nine designs.

\subsection{Robustness and coverage--width trade-offs}

\paragraph{The coverage improvement extends to different pretrained models while interval widths vary across models.} We evaluate TabCUQ with TabPFN 3 and TabFM using the same experimental design and perturbation budget as a matched TabPFN 2.6 comparison; see Table~\ref{tab:model:choice}. Across the four univariate designs, TabCUQ achieves edge-average coverage of \(0.986\)--\(1.000\) with TabPFN 3 and \(0.971\)--\(0.998\) with TabFM, compared with \(0.854\)--\(0.936\) and \(0.784\)--\(0.900\) for the corresponding Wild bootstrap intervals. Similar improvements are observed in the multivariate experiments, where TabCUQ achieves tail-average coverage of \(0.996\)--\(1.000\) for TabPFN 3 and \(0.998\)--\(1.000\) for TabFM. The relative widths remain substantial and depend on the pretrained model, ranging from \(1.7\) to \(2.4\) times the corresponding OBA benchmark across these comparisons. 

\begin{table}[htb!]
\centering
\small
\setlength{\tabcolsep}{4pt}
\caption{\label{tab:model:choice} Sensitivity to the choice of foundation models. Entries give ranges across all the simulation settings. Panel A concerns the eight univariate edge queries in U1--U4; Panel B concerns the 20 queries in the $90$--$99\%$ Mahalanobis stratum. Relative width is the mean pointwise TabCUQ/OBA width ratio, using each model's own Monte Carlo bias and standard deviation to construct its OBA reference. All three models are evaluated on the same nine multivariate settings: M1 at $p\in\{5,10\}$ and $\rho\in\{0,0.5\}$, M2--M4 at $p=5$ and $\rho=0$, and M9 at $p=2$ and $\rho\in\{0,0.5\}$.}
\label{tab:models}
\begin{tabular}{@{}lcrrr@{}}
\toprule
Model & Designs & \shortstack{Wild bootstrap\\coverage} & \shortstack{TabCUQ\\coverage} & \shortstack{TabCUQ width\\/ OBA} \\
\midrule
\multicolumn{5}{@{}l}{\textit{A. Univariate edge queries}}\\
TabPFN 2.6 & 4 & $0.738$--$0.910$ & $0.966$--$1.000$ & $1.773$--$2.082$ \\
TabPFN 3 & 4 & $0.854$--$0.936$ & $0.986$--$1.000$ & $1.967$--$2.419$ \\
TabFM & 4 & $0.784$--$0.900$ & $0.971$--$0.998$ & $1.722$--$2.164$ \\
\midrule
\multicolumn{5}{@{}l}{\textit{B. Multivariate tail queries}}\\
TabPFN 2.6 & 9 & $0.633$--$0.851$ & $0.981$--$0.998$ & $1.651$--$2.072$ \\
TabPFN 3 & 9 & $0.852$--$0.915$ & $0.996$--$1.000$ & $2.044$--$2.291$ \\
TabFM & 9 & $0.861$--$0.940$ & $0.998$--$1.000$ & $2.131$--$2.442$ \\
\bottomrule
\end{tabular}
\end{table}

\paragraph{Increasing the context size reduces interval width but does not necessarily resolve Wild bootstrap undercoverage.} We examine context sizes ranging from \(n=100\) to \(n=3000\) in the four original univariate designs. In the linear design, increasing \(n\) from \(100\) to \(3000\) reduces the average TabCUQ interval width from \(3.86\) to \(1.31\), while coverage increases from \(0.979\) to \(0.999\). However, Wild bootstrap coverage remains at \(0.845\) even when \(n=3000\). The OBA interval also becomes narrower, with its mean width decreasing from \(2.45\) to \(0.66\). Thus, although larger contexts improve absolute precision, they do not necessarily eliminate the need for bias protection or reduce the relative width cost of TabCUQ. Additional sample-size results are reported in Appendix~E.


\paragraph{The perturbation budget moderately affects the empirical bias cushion.} Because the cushion is defined through the maximum absolute displacement among perturbed estimates, adding more perturbations cannot decrease this maximum when the draws are nested. In the four original univariate designs, increasing the number of perturbations from \(100\) to \(500\) increases the average wild-bootstrap displacement cushion by \(8.5\%\)--\(14.4\%\). Thus, the perturbation budget moderately affects the interval being constructed rather than merely the numerical precision of its computation. Additional results are reported in Appendix~E.


\section{Real-World Studies}

\subsection{Data sets and study design}

We evaluate TabCUQ on seven real-world regression tasks spanning socioeconomic, environmental, behavioural, protein, and single-cell data. These tasks represent different prediction settings. Census income has a weak predictive signal and a heavy-tailed outcome, whereas tree-cover prediction has a stronger but highly nonlinear relationship between the covariates and outcome. Job Corps and Jester Jokes provide moderate- and high-dimensional regression settings. The two protein-stability tasks evaluate generalization to mutations on proteins largely absent from the training pool, with close sequence homologues removed. CITE-seq examines uncertainty quantification using \(2{,}000\) gene-expression features. We consider the task of CATE inference on the Jester Jokes dataset while implementing conditional-mean inference on all the other datasets. Detailed dataset descriptions and preprocessing procedures are provided in Appendix~I.

\begin{table}[!htbp]
\centering
\small
\caption{\label{tab:datasets} Basic information of the real-world datasets included in our studies.}
\begin{tabular}{lrrrr}
\toprule
Dataset & Pool & Evaluation & Input $p$ & Outcome \\
\midrule
Census & 11,800 & 200 & 2 & Income \\
Tree cover & 11,800 & 200 & 4 & Canopy cover \\
Job Corps & 10,014 & 200 & 23 & Earnings \\
Jester Jokes & 48,245 & 200 & 110 & Rating \\
Protein engineering $\Delta\Delta G$ & 8,754 & 461 & 145 & Gibbs free energy changes ($\Delta\Delta G$) \\
Protein engineering $\Delta T_m$ & 4,346 & 571 & 155 & Melting temperature changes ($\Delta T_m$) \\
CITE-seq CD71 & 30,469 & 200 & 2,000 & CD71 (CLR) \\
\bottomrule
\end{tabular}
\end{table}

Unlike in simulations, the population conditional mean or CATE is unknown in real data. We therefore evaluate confidence intervals relative to a fixed, data-derived reference function. For each dataset, we reserve an evaluation set and fit the reference function using the remaining observations. The reference model is selected by five-fold cross-validation from a collection of regression methods, including random forests, XGBoost, LightGBM, multilayer perceptrons, ridge regression, and a constant predictor. {A simple Cell-mean predictor is additionally included for the Census dataset with discrete covariates only.} The reference function is fitted using a substantially larger sample than the contexts supplied to the foundation model and is treated as fixed during evaluation. Consequently, the reported coverage concerns the estimated reference function rather than the unknown population conditional-mean or CATE functions.

We repeatedly sample contexts of size \(n_{\mathrm{ctx}}\in\{200,300,500,1000\}\) without replacement from the reference pool. For each sampled context, we obtain a point estimate and confidence interval at the same fixed evaluation queries. The reference function and evaluation queries remain unchanged across repetitions. We measure reference coverage by the proportion of intervals containing the large-sample reference function and summarize the interval widths across the repeated contexts. The main analyzes use \(200\) held-out evaluation points for all tasks except protein-stability prediction, while the two protein tasks use the evaluation sets of \(461\) and \(571\) mutations prespecified by the data providers. {For each sampled context, we generate 100 wild-bootstrap perturbations of the observed outcomes for TabCUQ.} Additional details on the setting and evaluation strategy are reported in Appendix~J.

\subsection{Results}

\begin{figure}[htb!]
    \centering
    \includegraphics[
        width=\linewidth,
        height=0.8\textheight,
        keepaspectratio
    ]{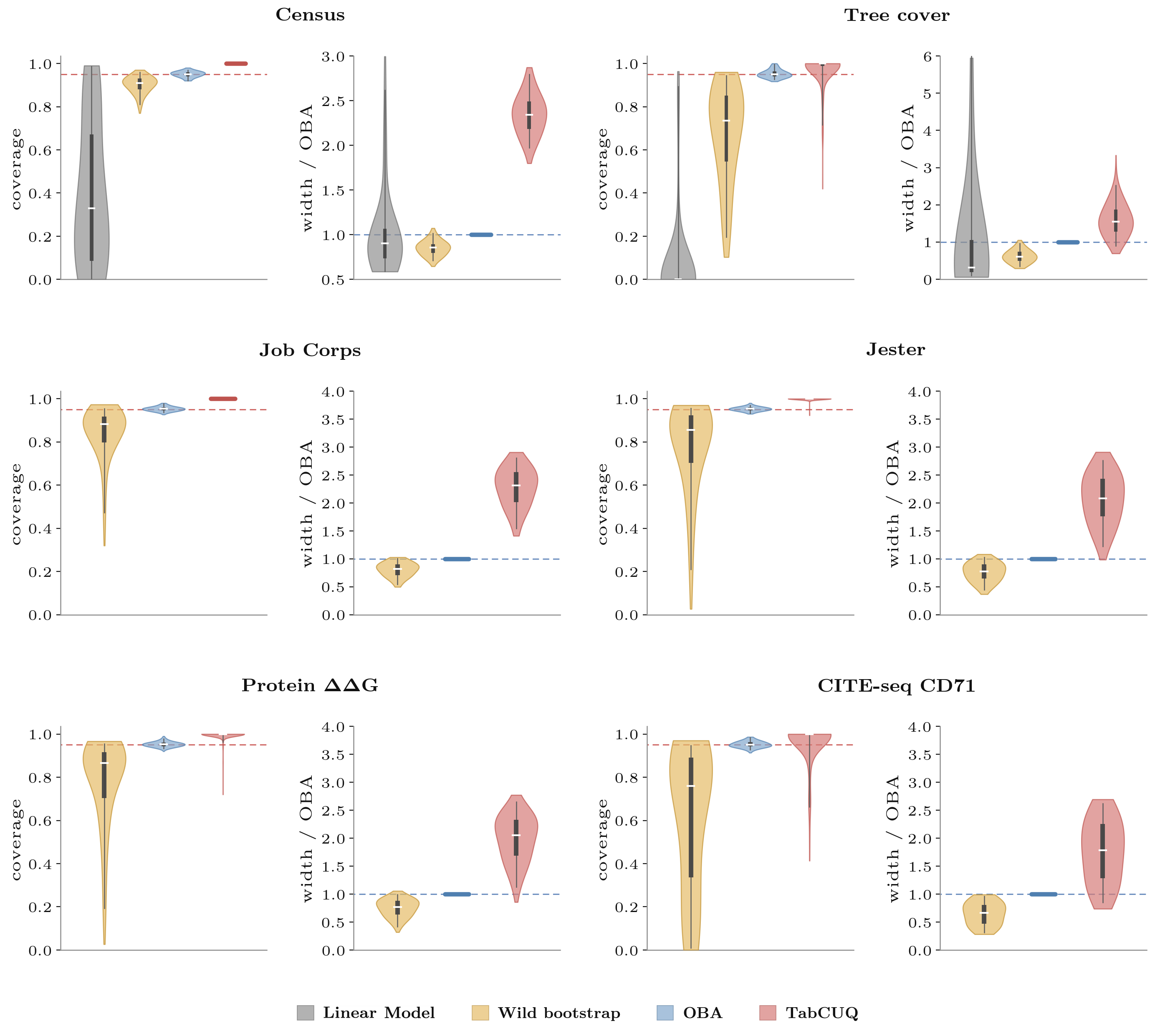}
    \caption{\label{fig:real} Pointwise reference coverage and relative interval width across six real-world tasks, with context size $n=1{,}000$. For each task, the left panel shows the distribution across evaluation queries of reference coverage. The right panel shows the distribution of mean interval width over those contexts divided by the OBA width at the same query. Wild bootstrap uses the estimated standard error without the empirical bias cushion; TabCUQ adds the empirical bias cushion from the same wild-bootstrap perturbations. Linear model denotes ordinary least-squares intervals, included only for the Census and Tree cover examples. OBA stands for the oracle benchmark. Inner boxes indicate medians and interquartile ranges; horizontal bars indicate identical values across queries. Dashed lines mark $0.95$ coverage and a width ratio of one.}
    \label{fig:realdata-pointwise}
\end{figure}

\paragraph{TabCUQ substantially improves reference coverage with interval widths comparable in scale to the oracle benchmark.} At \(n_{\mathrm{ctx}}=1000\), Wild bootstrap intervals achieve average reference coverage between \(0.614\) and \(0.842\) across the seven tasks, whereas intervals with the wild-bootstrap displacement cushion achieve coverage between \(0.966\) and \(1.000\). The mean pointwise TabCUQ/OBA width ratios range from $1.62$ to $2.27$, indicating that the coverage improvements are achieved with interval widths of the same order as those of the oracle bias-aware benchmark. The improvement is observed in various inferential tasks. For example, coverage increases from \(0.649\) to \(0.997\) for Census income and from \(0.682\) to \(0.973\) for tree-cover prediction. Similar improvements occur in the biological application examples with higher dimensional predictors. In specific, the coverage increases from \(0.779\) to \(0.997\) for protein \(\Delta\Delta G\), from \(0.658\) to \(0.967\) for protein \(\Delta T_m\), and from \(0.614\) to \(0.966\) for CITE-seq. These results show that the improvement is not confined to weak-signal problems, as substantial gains are also observed in nonlinear regression and molecular prediction tasks. Figure \ref{fig:real} presents pointwise coverage and relative-width comparisons for six of the tasks, with complete results for all seven reported in Appendix J.


\paragraph{Increasing the context size does not consistently resolve Wild bootstrap undercoverage.} In the context-size sensitivity experiments, Wild bootstrap reference coverage decreases from \(0.828\) to \(0.587\) for Census and from \(0.855\) to \(0.761\) for Jester as \(n_{\mathrm{ctx}}\) increases from \(100\) to \(2000\). Tree cover shows a different pattern, with coverage increasing from \(0.503\) to \(0.689\), although it remains below the nominal level. Undercoverage also persists in the protein-stability tasks, where Wild bootstrap coverage ranges from \(0.779\) to \(0.812\) for \(\Delta\Delta G\) and from \(0.658\) to \(0.689\) for \(\Delta T_m\) across context sizes of \(100\)--\(1000\). In a separate CITE-seq context-size analysis using \(500\) principal components, Wild bootstrap coverage increases from \(0.703\) to \(0.760\) but remains below nominal. This analysis uses a different feature representation and reference function from the main CITE-seq experiment. In contrast, TabCUQ intervals maintain reference coverage close to or above \(0.95\) across most context sizes and datasets. The contrasting trends across datasets indicate that increasing the number of context observations does not necessarily lead to better calibrated intervals. The complete results are provided in Appendix~J.

\paragraph{Favorable average reference coverage can still conceal substantial differences across individual queries.} This is particularly evident in the main CITE-seq experiment using \(1{,}000\) gene-expression features. Although the TabCUQ intervals achieve average reference coverage of \(0.966\), approximately \(10\%\) of evaluation points have pointwise coverage below \(0.90\). The Job Corps and Jester analyses also show considerable variation in interval width across queries. These results suggest that the improvement in reference coverage is not uniform across evaluation points, and that favorable dataset-level averages should not be interpreted as evidence of adequate coverage at every query.

\FloatBarrier

\section*{AI use statement}
We used generative AI tools to improve the clarity and wording of the manuscript and to assist in searching for relevant literature. The authors take full responsibility for the final content of the paper, including its scientific claims and references.

\bibliographystyle{plainnat}
\bibliography{references}

@book{wasserman2006all,
  author    = {Wasserman, Larry},
  title     = {All of Nonparametric Statistics},
  year      = {2006},
  publisher = {Springer},
  series    = {Springer Texts in Statistics},
  doi       = {10.1007/0-387-30623-4}
}

@article{xu2024improving,
  title   = {Improving the Prediction of Protein Stability Changes upon Mutations by Geometric Learning and a Pre-Training Strategy},
  author  = {Xu, Yunxin and Liu, Di and Gong, Haipeng},
  journal = {Nature Computational Science},
  volume  = {4},
  pages   = {840--850},
  year    = {2024},
  doi     = {10.1038/s43588-024-00716-2}
}

@article{hollmann2025tabpfn,
  title   = {Accurate Predictions on Small Data with a Tabular Foundation Model},
  author  = {Hollmann, Noah and M{\"u}ller, Samuel and Purucker, Lennart and Krishnakumar, Arjun and K{\"o}rfer, Max and Hoo, Shi Bin and Schirrmeister, Robin Tibor and Hutter, Frank and others},
  journal = {Nature},
  volume  = {637},
  pages   = {319--326},
  year    = {2025},
  doi     = {10.1038/s41586-024-08328-6}
}

@article{qu2025tabicl,
  title   = {{TabICL}: A Tabular Foundation Model for In-Context Learning on Large Data},
  author  = {Qu, Jingang and Holzm{\"u}ller, David and Varoquaux, Ga{\"e}l and Le Morvan, Marine},
  journal = {arXiv preprint arXiv:2502.05564},
  year    = {2025},
  url     = {https://arxiv.org/abs/2502.05564}
}

@misc{kong2026tabfm,
  title        = {Introducing {TabFM}: A Zero-Shot Foundation Model for Tabular Data},
  author       = {Kong, Weihao and Das, Abhimanyu},
  year         = {2026},
  howpublished = {Google Research Blog},
  url          = {https://research.google/blog/introducing-tabfm-a-zero-shot-foundation-model-for-tabular-data/},
  note         = {Accessed 2026-08-27}
}

@article{efron1979bootstrap,
  title   = {Bootstrap Methods: Another Look at the Jackknife},
  author  = {Efron, Bradley},
  journal = {The Annals of Statistics},
  volume  = {7},
  number  = {1},
  pages   = {1--26},
  year    = {1979},
  doi     = {10.1214/aos/1176344552}
}

@article{wu1986jackknife,
  title   = {Jackknife, Bootstrap and Other Resampling Methods in Regression Analysis},
  author  = {Wu, C. F. Jeff},
  journal = {The Annals of Statistics},
  volume  = {14},
  number  = {4},
  pages   = {1261--1295},
  year    = {1986},
  doi     = {10.1214/aos/1176350142}
}

@book{politis1999subsampling,
  title     = {Subsampling},
  author    = {Politis, Dimitris N. and Romano, Joseph P. and Wolf, Michael},
  publisher = {Springer},
  address   = {New York},
  year      = {1999},
  doi       = {10.1007/978-1-4612-1554-7}
}

@article{armstrong2016honest,
  title   = {Simple and Honest Confidence Intervals in Nonparametric Regression},
  author  = {Armstrong, Timothy B. and Koles{\'a}r, Michal},
  journal = {arXiv preprint arXiv:1606.01200},
  year    = {2016},
  url     = {https://arxiv.org/abs/1606.01200}
}

@article{lalonde1986evaluating,
  author  = {LaLonde, Robert J.},
  title   = {Evaluating the Econometric Evaluations of Training Programs with Experimental Data},
  journal = {The American Economic Review},
  year    = {1986},
  volume  = {76},
  number  = {4},
  pages   = {604--620},
  url     = {https://www.jstor.org/stable/1806062}
}

@article{dehejia1999causal,
  author  = {Dehejia, Rajeev H. and Wahba, Sadek},
  title   = {Causal Effects in Nonexperimental Studies: Reevaluating the Evaluation of Training Programs},
  journal = {Journal of the American Statistical Association},
  year    = {1999},
  volume  = {94},
  number  = {448},
  pages   = {1053--1062},
  doi     = {10.1080/01621459.1999.10473858}
}

@article{nagler2025uncertainty,
  author  = {Nagler, Thomas and R{\"u}gamer, David},
  title   = {Uncertainty Quantification for Prior-Data Fitted Networks using Martingale Posteriors},
  journal = {arXiv preprint arXiv:2505.11325},
  year    = {2025},
  doi     = {10.48550/arXiv.2505.11325},
  url     = {https://arxiv.org/abs/2505.11325}
}

@inproceedings{ng2026tabmgp,
  author    = {Ng, Kenyon and Fong, Edwin and Frazier, David T. and Knoblauch, Jeremias and Wei, Susan},
  title     = {{TabMGP}: Martingale Posterior with {TabPFN}},
  booktitle = {International Conference on Machine Learning},
  year      = {2026},
  url       = {https://arxiv.org/abs/2510.25154}
}

@article{shields2021bayesian,
  author = {Shields, Benjamin J. and Stevens, Jason and Li, Jun
            and Parasram, Marvin and Damani, Farhan
            and Martinez Alvarado, Jesus I. and Janey, Jacob M.
            and Adams, Ryan P. and Doyle, Abigail G.},
  title = {Bayesian reaction optimization as a tool for chemical synthesis},
  journal = {Nature},
  volume = {590},
  pages = {89--96},
  year = {2021},
  doi = {10.1038/s41586-021-03213-y}
}

@article{hickman2022constraints,
  author = {Hickman, Riley J. and Aldeghi, Matteo
            and H{\"a}se, Florian and Aspuru-Guzik, Al{\'a}n},
  title = {Bayesian optimization with known experimental and
           design constraints for chemistry applications},
  journal = {Digital Discovery},
  volume = {1},
  pages = {732--744},
  year = {2022},
  doi = {10.1039/D2DD00028H}
}

@inproceedings{muller2022transformers,
  title     = {Transformers Can Do {Bayesian} Inference},
  author    = {M{\"u}ller, Samuel and Hollmann, Noah and
               Pineda Arango, Sebastian and Grabocka, Josif and
               Hutter, Frank},
  booktitle = {International Conference on Learning Representations},
  year      = {2022},
  url       = {https://openreview.net/forum?id=KSugKcbNf9}
}

@article{lei2021conformal,
  title   = {Conformal Inference of Counterfactuals and Individual
             Treatment Effects},
  author  = {Lei, Lihua and Cand{\`e}s, Emmanuel J.},
  journal = {Journal of the Royal Statistical Society:
             Series B (Statistical Methodology)},
  year    = {2021},
  volume  = {83},
  number  = {5},
  pages   = {911--938},
  doi     = {10.1111/rssb.12445},
  url     = {https://doi.org/10.1111/rssb.12445}
}

@article{park2026conformal,
  author  = {Park, Sungwoo and Park, Sunghee and Chang, Won},
  title   = {Conformal Prediction and Conditional Coverage for Tabular Foundation Models},
  journal = {arXiv preprint arXiv:2609.34887},
  year    = {2026},
  url     = {https://arxiv.org/abs/2609.34887}
}

@article{fortini2026uncertainty,
  author  = {Fortini, Sandra and Ng, Kenyon and Petrone, Sonia
             and Rousseau, Judith and Wei, Susan},
  title   = {Uncertainty Decomposition for {Bayes}-Filtered Transformers
             via {Bayesian} Predictive Inference},
  journal = {arXiv preprint arXiv:2602.04596},
  year    = {2026},
  url     = {https://arxiv.org/abs/2602.04596}
}

@book{hastie2009elements,
  author    = {Hastie, Trevor and Tibshirani, Robert and Friedman, Jerome},
  title     = {The Elements of Statistical Learning: Data Mining, Inference, and Prediction},
  edition   = {2},
  publisher = {Springer},
  year      = {2009},
  doi       = {10.1007/978-0-387-84858-7}
}

@article{hines2022demystifying,
  author  = {Hines, Oliver and Dukes, Oliver and Diaz-Ordaz, Karla
             and Vansteelandt, Stijn},
  title   = {Demystifying Statistical Learning Based on Efficient
             Influence Functions},
  journal = {The American Statistician},
  year    = {2022},
  volume  = {76},
  number  = {3},
  pages   = {292--304},
  doi     = {10.1080/00031305.2021.2021984}
}

@article{kennedy2023towards,
  author  = {Kennedy, Edward H.},
  title   = {Towards Optimal Doubly Robust Estimation of
             Heterogeneous Causal Effects},
  journal = {Electronic Journal of Statistics},
  year    = {2023},
  volume  = {17},
  number  = {2},
  pages   = {3008--3049},
  doi     = {10.1214/23-EJS2157}
}

@book{hernan2020causal,
  author    = {Hern{\'a}n, Miguel A. and Robins, James M.},
  title     = {Causal Inference: {What If}},
  year      = {2020},
  publisher = {Chapman \& Hall/CRC},
  address   = {Boca Raton},
  url       = {https://miguelhernan.org/whatifbook}
}

@article{athey2016recursive,
  author  = {Athey, Susan and Imbens, Guido},
  title   = {Recursive partitioning for heterogeneous causal effects},
  journal = {Proceedings of the National Academy of Sciences},
  year    = {2016},
  volume  = {113},
  number  = {27},
  pages   = {7353--7360},
  doi     = {10.1073/pnas.1510489113}
}

@article{kunzel2019metalearners,
  author  = {K{\"u}nzel, S{\"o}ren R. and Sekhon, Jasjeet S. and Bickel, Peter J. and Yu, Bin},
  title   = {Metalearners for estimating heterogeneous treatment effects using machine learning},
  journal = {Proceedings of the National Academy of Sciences},
  year    = {2019},
  volume  = {116},
  number  = {10},
  pages   = {4156--4165},
  doi     = {10.1073/pnas.1804597116}
}

@book{wasserman2004all,
  author    = {Wasserman, Larry},
  title     = {All of Statistics: A Concise Course in Statistical Inference},
  publisher = {Springer},
  year      = {2004},
  doi       = {10.1007/978-0-387-21736-9}
}

@article{mammen1993bootstrap,
    author  = {Mammen, Enno},
    title   = {Bootstrap and Wild Bootstrap for High Dimensional Linear Models},
    journal = {The Annals of Statistics},
    year    = {1993},
    volume  = {21},
    number  = {1},
    pages   = {255--285},
    doi     = {10.1214/aos/1176349025}
}

@article{kline2012score,
    author  = {Kline, Patrick and Santos, Andres},
    title   = {A Score Based Approach to Wild Bootstrap Inference},
    journal = {Journal of Econometric Methods},
    year    = {2012},
    volume  = {1},
    number  = {1},
    pages   = {23--41},
    doi     = {10.1515/2156-6674.1006}
}

@article{shafer2008tutorial,
  title   = {A Tutorial on Conformal Prediction},
  author  = {Shafer, Glenn and Vovk, Vladimir},
  journal = {Journal of Machine Learning Research},
  volume  = {9},
  number  = {12},
  pages   = {371--421},
  year    = {2008},
  url     = {https://jmlr.org/papers/v9/shafer08a.html}
}

@article{lei2018distribution,
  title   = {Distribution-Free Predictive Inference for Regression},
  author  = {Lei, Jing and G'Sell, Max and Rinaldo, Alessandro
             and Tibshirani, Ryan J. and Wasserman, Larry},
  journal = {Journal of the American Statistical Association},
  volume  = {113},
  number  = {523},
  pages   = {1094--1111},
  year    = {2018},
  doi     = {10.1080/01621459.2017.1307116}
}

@article{barber2021limits,
  title   = {The Limits of Distribution-Free Conditional Predictive Inference},
  author  = {Barber, Rina Foygel and Cand{\`e}s, Emmanuel J.
             and Ramdas, Aaditya and Tibshirani, Ryan J.},
  journal = {Information and Inference: A Journal of the IMA},
  volume  = {10},
  number  = {2},
  pages   = {455--482},
  year    = {2021},
  doi     = {10.1093/imaiai/iaaa017}
}

@book{tsybakov2009,
    author    = {Tsybakov, Alexandre B.},
    title     = {Introduction to Nonparametric Estimation},
    series    = {Springer Series in Statistics},
    publisher = {Springer},
    address   = {New York},
    year      = {2009},
    doi       = {10.1007/b13794},
    url       = {https://doi.org/10.1007/b13794}
}

\end{document}